\documentclass[sn-mathphys-num]{sn-jnl}

\usepackage{graphicx}%
\usepackage{multirow}%
\usepackage{amsmath,amssymb,amsfonts}%
\usepackage{amsthm}%
\usepackage{mathrsfs}%
\usepackage[title]{appendix}%
\usepackage{xcolor}%
\usepackage{lscape}
\usepackage{textcomp}%
\usepackage{manyfoot}%
\usepackage{booktabs}%
\usepackage{algorithm}%
\usepackage{algorithmicx}%
\usepackage{algpseudocode}%
\usepackage{listings}%

\theoremstyle{thmstyleone}%
\theoremstyle{thmstyletwo}%

\theoremstyle{thmstylethree}%

\begin{document}

\title[Article Title]{ReLU-KAN: New Kolmogorov-Arnold Networks that Only Need Matrix Addition, Dot Multiplication, and ReLU}


\author[1]{\fnm{Qi} \sur{Qiu}}\email{qiuqi@stu.usc.edu.cn}

\author*[1]{\fnm{Tao} \sur{Zhu}}\email{tzhu@usc.edu.cn}

\author[2]{\fnm{Helin} \sur{Gong}}\email{gonghelin@sjtu.edu.cn}

\author[3]{\fnm{Liming} \sur{Chen}}\email{LimingChen0922@dlut.edu.cn}

\author[4]{\fnm{Huansheng} \sur{Ning}}\email{ninghuansheng@ustb.edu.cn}

\affil*[1]{\orgdiv{School of Computer Science}, \orgname{University of South China}, \orgaddress{\city{Hengyang}, \postcode{420000}, \state{Hunan}, \country{China}}}

\affil[2]{\orgdiv{Paris Elite Institute of Technology}, \orgname{Shanghai Jiao Tong University}, \orgaddress{\city{Shanghai}, \postcode{200000}, \country{China}}}

\affil[3]{\orgdiv{School of Computer Science and Technology}, \orgname{University of Science and Technology Beijing}, \orgaddress{\city{Beijing}, \postcode{100083} \country{China}}}

\affil[4]{\orgdiv{School of Computer and Communication Engineering}, \orgname{Dalian University of Technology}, \orgaddress{\city{Dalian}, \postcode{116000}, \state{Liaoning}, \country{China}}}


\abstract{We propose a novel KAN architecture that replaces the original basis function (B-spline) with a new one more suitable for parallel computation. The proposed basis function is composed solely of matrix addition, dot product, and ReLU activation, enabling efficient GPU parallelization. Unlike the static B-splines, novel basis function incorporates two trainable hyperparameters that allow it to dynamically adapt its shape and position to the specific fitting task. This adaptive capability gives ReLU-KAN a significant advantage in modeling complex functions. Experimental results on a four-layer network show a 20-fold speedup in backpropagation and a the accuracy is improved by two to three orders of magnitude compared to the original KAN. Notably, ReLU-KAN preserves the original model's ability to avoid catastrophic forgetting. The source code is available at https://github.com/quiqi/relu\_kan}


\maketitle

\section{Introduction}\label{sec1}

Kolmogorov-Arnold Networks (KANs) \cite{1liu2024kan} have recently garnered significant interest due to their exceptional performance and novel architecture \cite{2abueidda2024deepokan,2genet2024tkan,1samadi2024smooth}. Researchers have rapidly adopted KANs for tackling diverse problems \cite{3bozorgasl2024wav,3vaca2024kolmogorov,2xu2024kolmogorov} or combine KAN with existing network structures\cite{xiugaibresson2024kagnns,xiugaikiamari2024gkan}. 

The KANs was designed to overcome the limitation of MLPs, where each layer can only perform linear transformations and requires stacking multiple layers with activation functions to fit complex non-linear relationships. In contrast, KANs can fit complex non-linear relationships with a single layer, resulting in a substantial improvement in parameter utilization efficiency.

However, the recursive nature of B-splines and their irregular control point distribution can lead to computational inefficiencies. Our proposed simplified basis function, expressed as a combination of ReLU activations, offers a more straightforward and computationally efficient representation. By reformulating the basis function computation as matrix operations and leveraging convolution, we fully exploit the parallel processing capabilities of GPUs. Moreover, we found in practice that because B-spline functions are unable to change their position and shape during training, it is difficult to fit those functions that change in the domain with high frequency. While increasing the number of spline functions can enhance fitting capabilities, it also introduces challenges such as a more complex training process, an increased number of parameters, and longer training times.

In this paper, we present an innovative spline function as a replacement for the B-spline. This new function is constructed entirely from matrix multiplications, additions, and ReLU activations, as described in Eq~\ref{r_i}. By treating $e_i$ and $s_i$ as trainable parameters, we endow the basis function with greater plasticity.

\begin{equation}
R_i(x) = [\mathrm{ReLU}(e_i - x)\times \mathrm{ReLU}(x-s_i)]^2 \times 16 / (b_i - a_i)^4 \label{r_i}
\end{equation}

We evaluated ReLU-KAN's performance on a set of functions used in the original KAN paper. Compared to KAN, ReLU-KAN demonstrates significant improvements in training speed, convergence stability, and fitting accuracy, particularly for larger network architectures. Notably, ReLU-KAN inherits most of KAN's key properties, including hyperparameters like the number of grids and its ability to prevent catastrophic forgetting. In the existing experiments, ReLU-KAN is 5 to 20 times faster than KAN in training, and the accuracy of ReLU-KAN is 1-3 orders of magnitude higher than KAN.

In summary, this paper presents a novel KAN architecture, ReLU-KAN, which significantly improves the training efficiency and accuracy of KANs. Our key contributions include: 
\begin{enumerate}
    \item Simplified basis function based on ReLU activations;
    \item Trainable hyperparameters within the basis function for improved adaptation to complex functions.
    \item Efficient matrix-based operations for GPU acceleration;
    \item A convolutional implementation that seamlessly integrates with existing deep learning frameworks.
\end{enumerate}
future work will explore the application of ReLU-KAN to more complex tasks and investigate the potential of combining it with other neural network architectures.

The remaining sections of this paper delve deeper into the details of our contributions. In Section~\ref{sec:2}, we introduce KANs and their relationship to Multi-Layer Perceptrons (MLPs). Section~\ref{sec:3} details the ReLU-KAN architecture, and provided its PyTorch implementation. Finally, Section~\ref{sec:4} presents comprehensive experiments that evaluate ReLU-KAN's performance against traditional KANs. We demonstrate ReLU-KAN's significant advantages in training speed, convergence stability, and fitting accuracy, particularly for larger networks. Additionally, we confirm ReLU-KAN's ability to prevent catastrophic forgetting.

\section{Related Works}
\label{sec:2}
This section provides an overview of Kolmogorov-Arnold Networks (KANs). In the first part of section, KAN is regarded as an extension of MLPs. At the end of Subsection \ref{sec:2.1}, a new idea of constructing KAN-like structures is proposed. Next, the application of B-splines in KAN is discussed to provide the basis for discussion in the methods section. At the end of the section, we present the state of the art in KANs.

\subsection{Introduce Kolmogorov-Arnold Networks as an Extension of MLPs} \label{sec:2.1}
\begin{figure}[htp]
  \centering
  \includegraphics[width=0.7\columnwidth]{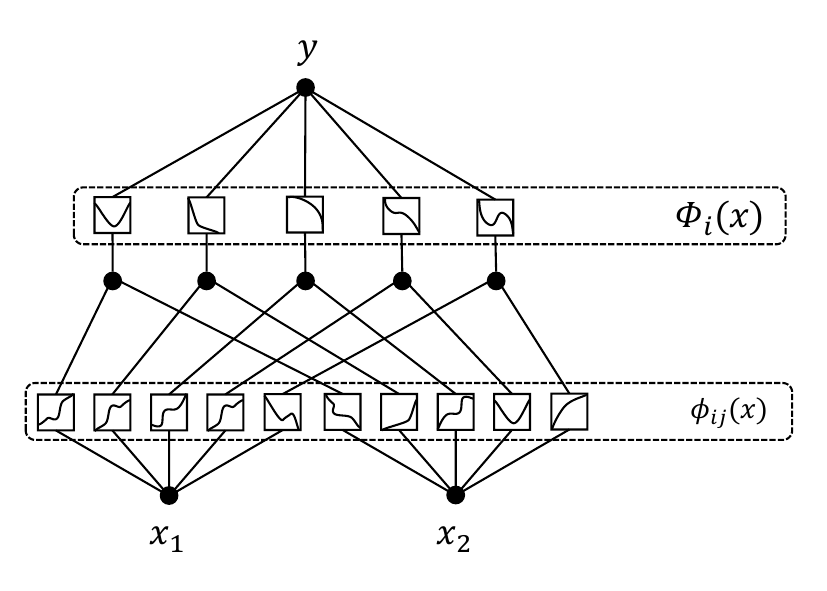}
  \caption{Kolmogorov-Arnold representation theorem can be presented a two-layer structure}
  \label{fig:fig_kan}
\end{figure}

The Kolmogorov-Arnold representation theorem confirms that a high-dimensional function can be represented as a composition of a finite number of one-dimensional functions as Eq~\ref{eq2} \cite{schmidt2021kolmogorov}.
\begin{equation} \label{eq2}
    f(\boldsymbol{x}) = \sum_{i=1}^{2n+1}\Phi_i(\sum_{j=1}^{n}\phi_{i,j}(x_j))
\end{equation}
where $\phi_{i,j}$ is called the inner function and $\Phi_i$ is called the outer function. Building upon the theorem's mathematical framework, the Kolmogorov-Arnold representation theorem can be presented a two-layer structure Fig~\ref{fig:fig_kan}.
Let's consider a KAN where the input vector, denoted by $\boldsymbol{x}$, has a length of $n$, and the output $y$, The Eq~\ref{eq3} describes the Fig~\ref{fig:fig_kan}
\begin{equation} \label{eq3}
    y = 
\begin{pmatrix}
	\Phi(\cdot)_{1} \\
	\Phi(\cdot)_{2} \\
	\vdots \\
	\Phi(\cdot)_{2n+1}
\end{pmatrix}
(
\begin{pmatrix}
	\phi(\cdot)_{1,1} & \phi(\cdot)_{1,2} & \cdots & \phi(\cdot)_{1,2n+1} \\
	\phi(\cdot)_{2,1} & \phi(\cdot)_{2,2} & \cdots & \phi(\cdot)_{1,2n+1} \\
	\vdots & \vdots & \ddots & \vdots \\
	\phi(\cdot)_{n,1} & \phi(\cdot)_{n,2} & \cdots & \phi(\cdot)_{1,2n+1}
\end{pmatrix} \boldsymbol{x})
\end{equation}
In order to ensure the representation power of $\phi_{ij}$and $\Phi_{i}$, they are represented as linear combinations of multiple B-spline functions and a bias function as Eq~\ref{eq4}:
\begin{equation} \label{eq4}
    \phi(x) = w_b  x/(1+e^{-x}) + w_s \sum c_i B_i(x)
\end{equation}
where $B_i(x)$ is a B-spline function.

Assuming we define $\phi_{ij}(x_j) = w_{ij} x_j, \Phi_i(x) = \text{ReLU}(x)$, the Eq~\ref{eq3} can be regarded as a MLP. This MLP takes a n-dimensional input, reduces it to a one-dimensional output, and employs a single hidden layer containing $2n+1$. In this sense, KANs can be thought of as an extension of MLPs. The activation function plays a crucial role in MLPs because the $\phi_{ij}(x_j) = w_{ij} x_j$ lacks nonlinear fitting ability. But if $\phi_{ij}(x)$ is a nonlinear function, the activation function can be omitted.

We can extend the hidden layer architecture of KAN networks similar to multi-layer perceptrons (MLPs). Consequently, a hidden layer processing n inputs and generating m outputs can be expressed using Eq~\ref{eq5} after relaxing the constraint that the number of nodes must be $2n+1$ and disregarding the activation function $\Phi(\cdot)$

and the KAN can be represented as Eq~\ref{eq5}:

\begin{equation} \label{eq5}
    \text{KAN}_{hidden}(x) = 
\begin{pmatrix}
	\phi(\cdot)_{11} & \phi(\cdot)_{12} & \cdots & \phi(\cdot)_{1n} \\
	\phi(\cdot)_{21} & \phi(\cdot)_{22} & \cdots & \phi(\cdot)_{2n} \\
	\vdots & \vdots & \ddots & \vdots \\
	\phi(\cdot)_{m1} & \phi(\cdot)_{m2} & \cdots & \phi(\cdot)_{mn}
\end{pmatrix} \boldsymbol{x}
\end{equation}
we can construct more KAN-like structures based on Eq~\ref{eq5}, just by finding suitable nonlinear $\phi(x)$.

\subsection{B-splines Application in KAN}
B-spline is a complex function based on recursion definition. Due to the specialization and complexity of its definition, and considering that it is not tightly coupled with the central idea of KAN, we will not discuss it in depth here, but focus on its application in KAN. 

\begin{figure}[htp]
  \centering
  \includegraphics[width=1.0\columnwidth]{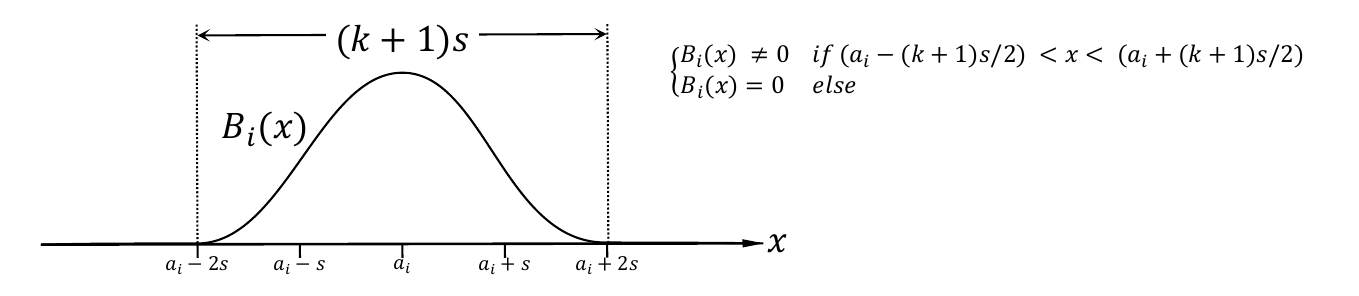}
  \caption{The $i^{th}$ B-spline.}
  \label{fig:fig1}
\end{figure}

A set of B-spline functions denoted as $\boldsymbol{B}=\{B_1(a_1, k, s, x), B_2(a_2, k, s, x), \dots,$ $ B_n(a_n, k, s, x)\}$ are used as basis functions to represent any unary function on a finite domain. These B-spline functions share the same shape but have different positions.  Each term $B_i(a_i, k, x)$ is a bell-shaped function, and $a_i$, $k$, and $s$ are the hyper-parameters of $B_i$ \cite{eilers1996flexible,chaudhuri2021b}. The $a_i$ is used to control the position of the symmetry axis, $k$ determines the range of the non-zero region, and $s$ is the unit interval. The $i$ spline $B_i$ is shown as Fig~\ref{fig:fig1}(Let's say $k=3$).

\begin{figure}
  \centering
  \includegraphics[width=1.0\columnwidth]{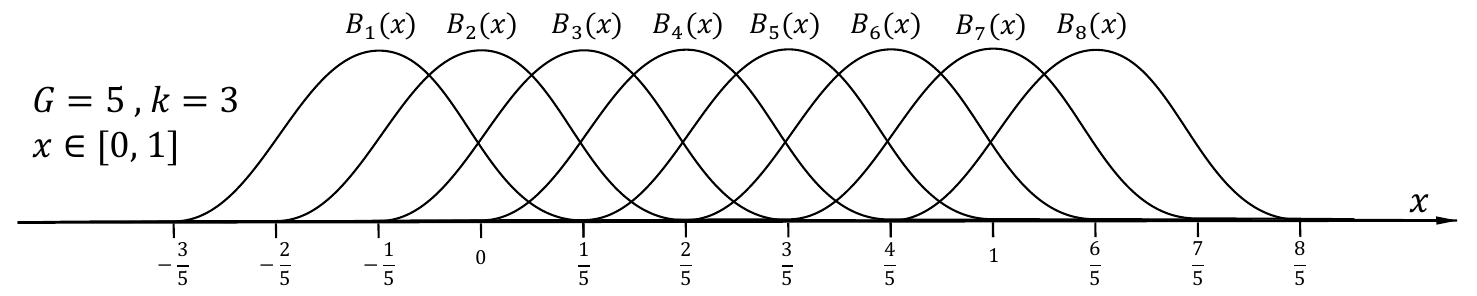}
  \caption{Appearance of $\boldsymbol{B}$ for the case of $G = 5$ and $k = 3$.}
  \label{fig:fig2}
\end{figure}

The hyperparameters of basis function set $\boldsymbol{B}$ depend on the number of grids, denoted by $G$. Specifically, when the domain of the function to be approximated is $x \in [0, 1]$, we have $n = G + k$ basis functions, step size: $s = 1/G$ and $a_i = \frac{2i+1-k}{2G}$.
Figure~\ref{fig:fig2} illustrates the appearance of $\boldsymbol{B}$ for the case of $G = 5$ and $k = 3$.

The function to be fitted $f(x)$ is expressed as Eq~\ref{eq4} in KAN. Using an optimization algorithm such as gradient descent to determine the values of $w_b, w_s$, and $\boldsymbol{c} = [c_1, c_2, \dots, c_n]$, we obtain $\phi(x)$ fitted using B-splines function.

Increasing the number of grids, $G$, leads to a greater number of trainable parameters, consequently enhancing the model's fitting ability. However, a larger k value strengthens the coupling between the B-spline functions, which can also improve fitting ability. As both $G$ and $k$ are effective hyper parameters for controlling the model's fitting capability, we and retain them within the ReLU-KAN architecture.

Although the solving process of spline function is difficult to characterize as matrix operation because of its complexity, and it is difficult to use the parallel ability of GPU \cite{de1972calculating}, from the above analysis, spline function is not necessary compared with KAN.

\subsection{Frontiers of KANs}
Since its inception, the Kolmogorov-Arnold Network (KAN) has garnered significant attention from both academia and industry due to its unique architecture and powerful expressive capabilities. Researchers have delved into KANs from both theoretical and applied perspectives, yielding fruitful results.

In terms of theoretical research, Altarabichi et al. conducted an in-depth study on alternative multivariate functions for KAN neurons, discovering that by restricting the input range, training stability can be significantly improved \cite{1altarabichi2024rethinkingfunctionneuronskans}. 
Yu et al. conducted a detailed comparison between KANs and traditional Multi-Layer Perceptrons (MLPs), revealing the strengths and weaknesses of the two models in various tasks \cite{2yu2024kanmlpfairercomparison}. 
Bodner et al. introduced Convolutional KANs by combining KANs with convolutional neural networks, effectively reducing the number of model parameters and providing new insights for optimizing neural network architectures \cite{3bodner2024convolutionalkolmogorovarnoldnetworks}. 
In terms of applications, Cheon demonstrated the effectiveness of KANs in computer vision tasks \cite{4cheon2024demonstratingefficacykolmogorovarnoldnetworks}. Li et al. combined KANs with U-Nets for medical image segmentation, achieving promising results \cite{6li2024ukanmakesstrongbackbone}. 
Abueidda et al. employed improved KANs to address mechanics-related problems, showcasing the potential of KANs in physical modeling \cite{2abueidda2024deepokan}.

Despite its broad application prospects, KANs still face several challenges. For instance, Shen et al. found that KANs are highly sensitive to noise, limiting their robustness in real-world applications \cite{7shen2024reducedeffectivenesskolmogorovarnoldnetworks}. 
Tran et al. delved into the limitations of KANs in classification tasks, providing directions for future research \cite{8tran2024exploringlimitationskolmogorovarnoldnetworks}. 
This paper will focus on hardware acceleration, tailoring optimizations to the characteristics of KANs to enhance their computational efficiency.

As an emerging neural network architecture, KANs hold immense potential. With ongoing research, KANs are expected to play a significant role in a wider range of domains.

\section{Methods}
\label{sec:3}
\subsection{ReLU-KAN}
We use the simpler function $R_i(x)$ to replace the B-spline function in KAN as the new basis function:
\begin{equation} \label{eq6}
    R_i(x) = [\text{ReLU}(e_i - x)\times \text{ReLU}(x-s_i)]^2 \times 16 / (e_i - s_i)^4
\end{equation}
where, $\text{ReLU}(x) = \text{max}(0, x)$.

It is easy to find that the maximum value of $\text{ReLU}(e_i - x)\times \text{ReLU}(x-s_i)$ is $\frac{(e_i - s_i)^2}{4}$ when $x=(e_i+s_i)/2$, so the maximum value of $[\text{ReLU}(e_i - x)\times \text{ReLU}(x-s_i)]^2$ is $\frac{(e_i - s_i)^4}{16}$, and $\frac{16}{(e_i - s_i)^4}$ is used as the normalization constant.

\begin{figure}
  \centering
  \includegraphics[width=1.0\columnwidth]{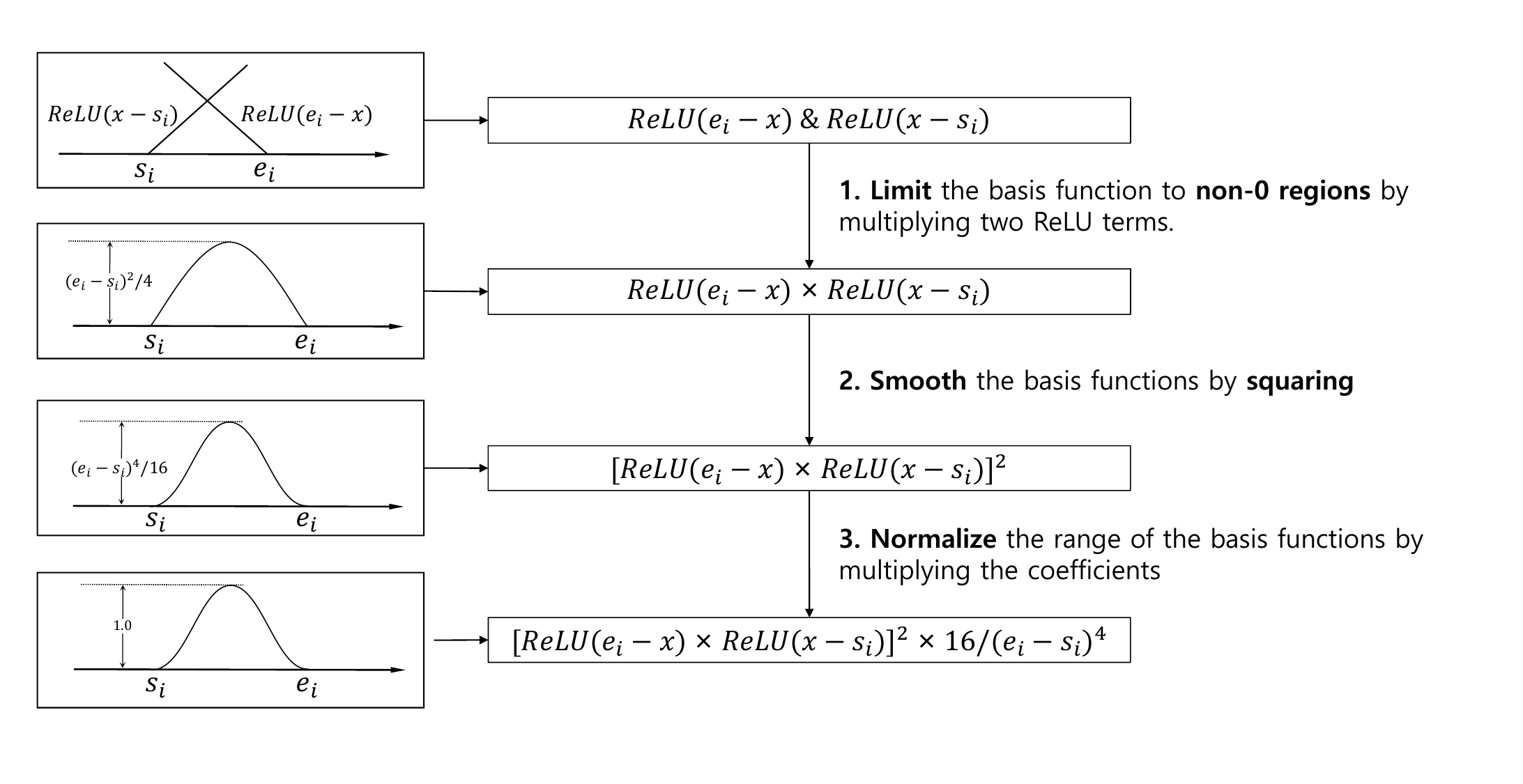}
  \caption{The construction of $R_i$}
  \label{fig:fig3}
\end{figure}

Like $B_i(x)$, $R_i(x)$ is also a unary bell-shaped function, which is nonzero at $x \in [s_i, e_i]$and zero at other intervals. The $\text{ReLU}(x)$ function is used to limit the range of nonzero values, and the squaring operation is used to increase the smoothness of the function. As Fig~\ref{fig:fig3}.

Multiple basis function $R_i$ can form the basis function set $\boldsymbol{R}= \{R_1 (x), R_2 (x), \dots, R_n (x) \}$, $\boldsymbol{R}$ inherited many properties of $\boldsymbol{B}$, It is again composed of $n$basis functions with the same shape but different positions, and the number of basis functions $n$ and $a_i, b_i$ are also determined by the number of grids $G$and the span parameter $k$.

A set of basis functions, denoted by $\boldsymbol{R}= \{R_1 (x), R_2 (x), \dots, R_n (x) \}$, can be constructed from multiple basis functions, $R_i$. And $\boldsymbol{R}$ inherits many properties from the $\boldsymbol{B}$,  $\boldsymbol{R}$ consists of $n$ basis functions with identical shapes but varying positions. The number of basis functions, $n$, along with the position parameters $a_i$ and $b_i$ are still determined by the number of grids, $G$, and the span parameter, $k$.

If we assume that the domain of the function to be fitted is $x \in [0, 1]$, the number of grids is $G$, and the span parameter is $k$, the number of spline functions is $n=G+k$. Where $s_i$ and $e_i$ are a set of trainable parameters, They denote the interval of the nonzero part of the basis function $R_i(x)$, Their initial values are set as follows: $s_i = \frac{i-k-1}{G}, e_i=\frac{i}{G}$. 

\begin{figure}
  \centering
  \includegraphics[width=1.0\columnwidth]{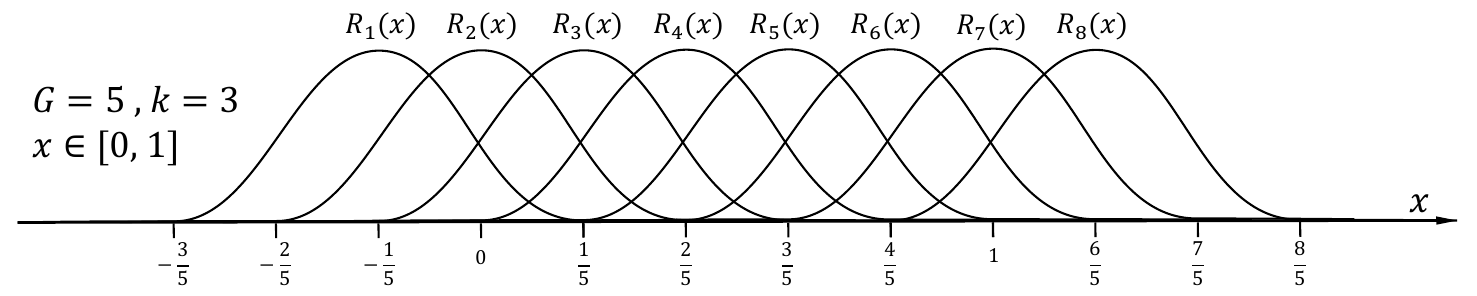}
  \caption{Appearance of $\boldsymbol{R}$ for the case of $G = 5$ and $k = 3$.}
  \label{fig:fig4}
\end{figure}

For example, the Fig~\ref{fig:fig4} shows a schematic representation of $\boldsymbol{R}$ for $G=5$ and $k=3$.

The ReLU-KAN layer can also be expressed by Eq(5), and the corresponding $\phi(x)$ of ReLU-KAN removes the bias function and is further simplified to Eq~\ref{eq7}.

\begin{equation} \label{eq7}
    \phi(x) = \sum_{i=1}^{G+k}{w_iR_i(x)}
\end{equation}

\begin{figure}
  \centering
  \includegraphics[width=0.7\columnwidth]{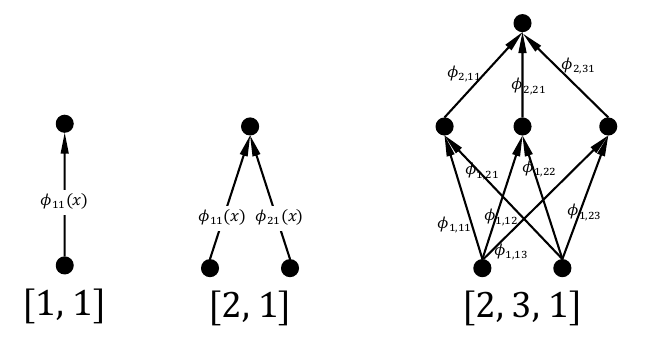}
  \caption{The multi-layer ReLU-KAN}
  \label{fig:fig5}
\end{figure}

The multi-layer ReLU-KAN can be represented as the Fig~\ref{fig:fig5} In the following expression, we use $[n_1, n_2, \dots, n_k]$to represent a ReLU-KAN of $k-1$ layers, and the $i$ layer takes as input the output of $i-1$ layers. Its input vector is of length $n_i$ and its output vector is of length $n_{i+1}$.

\subsection{Operational Optimization}
Consider the computation of ReLU KAN for a single layer. Given the hyper parameters $G$ and $k$, the number of inputs $n$ named $\boldsymbol{x} = [x^1, x^2, \dots, x^i, \dots, x^n]$, and the number of outputs $m$ named $\boldsymbol{y} = [y^1, y^2, \dots, y^c, \dots, y^m]$, we pre-compute the start matrix $S$, end matrix $E$ and $m$ weight matrix $\boldsymbol{W} = [W^1, W^2, \dots, W^c, \dots, W^m]$ as Eq~\ref{eq8}.

\begin{small}
    \begin{equation} \label{eq8}
    \tiny
S = 
\begin{pmatrix}
	s_{1,1} & s_{1,2} & \cdots & s_{1,G+k} \\
	s_{2,1} & s_{2,2} & \cdots & s_{2,G+k} \\
	\vdots & \vdots & \ddots & \vdots \\
	s_{n,1} & s_{n,2} & \cdots & s_{n,G+k}
\end{pmatrix} 
E = 
\begin{pmatrix}
	e_{1,1} & e_{1,2} & \cdots & e_{1,G+k} \\
	e_{2,1} & e_{2,2} & \cdots & e_{2,G+k} \\
	\vdots & \vdots & \ddots & \vdots \\
	e_{n,1} & e_{n,2} & \cdots & e_{n,G+k}
\end{pmatrix} 
W^c = 
\begin{pmatrix}
	w^c_{1,1} & w^c_{1,2} & \cdots & w^c_{1,G+k} \\
	w^c_{2,1} & w^c_{2,2} & \cdots & w^c_{2,G+k} \\
	\vdots & \vdots & \ddots & \vdots \\
	w^c_{n,1} & w^c_{n,2} & \cdots & w^c_{n,G+k}
\end{pmatrix}
\end{equation}
\end{small}

where, $s_{i,j} = \frac{j-k-1}{G}, e_{i,j}=\frac{j}{G}$ and $w^c_{i,j}$ is a random float number. As mentioned in the previous subsection, $S$ and $E$ are trainable.

When using Eq~\ref{eq6} as the basis function, we define a normalization constant $r=\frac{16G^4}{(k+1)^4}$, the computation of output $\boldsymbol{y} = [y^1, y^2, \dots, y^c, \dots, y^m]$ can be decomposed into the following matrix operation:
\begin{align}    
A &= \text{ReLU}(E - \boldsymbol{x}^T) \label{step1}\\
B &= \text{ReLU}(\boldsymbol{x}^T - S) \label{step2}\\
D &= r\times A \cdot B \label{step3}\\
F &= D \cdot D \label{step4}\\
\boldsymbol{y} &= \boldsymbol{W} \otimes F \label{step5}
\end{align}
where, $A, B, D$ and $F$ are all intermediate results. '$\cdot$' is the dot product operation. '$\otimes$' is the convolution operation commonly used in deep learning. Since $W^c$ and $F$ are equally large, Eq~\ref{step5} will output a vector length $m$. 

Eq~\ref{step1} to Eq~\ref{step4} used to compute all basis functions as Eq~\ref{eq6} in this ReLU-KAN layer, The result of these steps, $F$, can be described as Eq~\ref{eqF}
\begin{equation} \label{eqF}
    F = \begin{pmatrix}
        R_1(x_1) & R_2(x_1) & \cdots & R_{G+k}(x_1) \\
        R_1(x_2) & R_2(x_2) & \cdots & R_{G+k}(x_2) \\
        \vdots & \vdots & \ddots & \vdots \\
        R_1(x_n) & R_2(x_n) & \cdots & R_{G+k}(x_n)
\end{pmatrix}
\end{equation}

\begin{figure*}
  \centering
  \includegraphics[width=1\columnwidth]{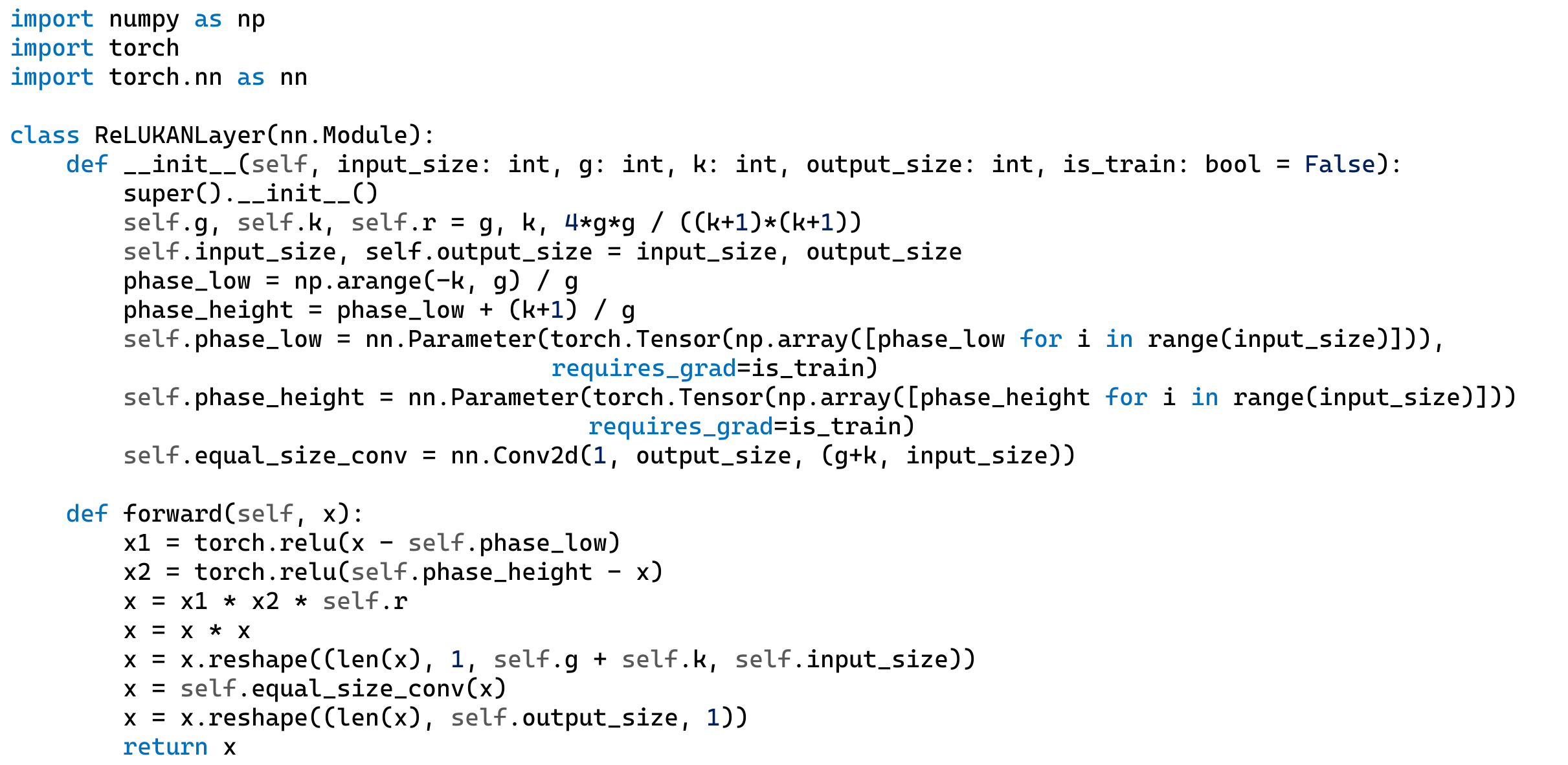}
  \caption{The ReLU-KAN Layer Code}
  \label{fig:code}
\end{figure*}

In the real code implementation, we can directly use the convolutional layer to implement the calculation of Eq~\ref{step5}. We give the Python code of the ReLU-KAN layer based on PyTorch as Fig~\ref{fig:code}. It is very simple code and does not need to take up too much space.

\section{Experiments} \label{sec:4}
The experimental evaluation is divided into three main parts. Firstly, the training speed of KAN and ReLU-KAN in GPU and CPU environments was compared. Second, we evaluate the fitting ability and convergence rate of the two models under the same parameter Settings, and we include ablation experiments that confirm the effect of trainable parameters in the basis functions in ReLU-KAN. Finally, ReLU-KAN is used to replicate the performance of KAN in the context of catastrophic forgetting.

\subsection{Training Speed Comparison}
We chose a function set of size 5 to compare KAN and ReLU-KAN training speeds. The parameters of KAN and ReLU-KAN are set as Table~\ref{tab:tab1}.
\begin{table}[]
    \centering
    \caption{Parameter Settings for Training Speed Comparison Experiments} \label{tab:tab1}
    \begin{tabular}{l|l|l}
    \hline
    Func. & KAN Setting & ReLU-KAN Setting \\ \hline
    $f_1: \text{sin}(\pi x)$ & width=[1, 1], G=5, k=3 & width=[1, 1], G=5, k=3 \\ \hline
    $f_2: \text{sin}(\pi x_1 + \pi x_2)$ & width=[2, 1], G=5, k=3 & width=[2, 1], G=5, k=3 \\ \hline
    $f_3: \text{arctan}(x_1 + x_1x_2 + x_2^2)$ & width=[2, 1, 1], G=5, k=3 & width=[2, 1, 1], G=5, k=3 \\ \hline
    $f_4: e^{\text{sin}(\pi x_1) + x_2^2}$ & width=[2, 5, 1], G=5, k=3 & width=[2, 5, 1], G=5, k=3 \\ \hline
    $f_5: e^{\text{sin}(x_1^2 + x_2^2)+\text{sin}(x_3^2 + x_4^2)}$ & width=[4, 4,  2, 1], G=10, k=3 & width=[4, 4, 2, 1], G=10, k=3 \\ \hline
\end{tabular}
\end{table}

The training process was conducted using the PyTorch framework. We employed the Adam optimizer for optimization and set the training set size to 1,000 samples. All models were trained for 500 iterations. The Table~\ref{tab:tab2} summarizes the training times for KAN and ReLU-KAN on both GPU and CPU environments, measured in seconds.

The Table~\ref{tab:tab2} compares two configurations of ReLU-KAN. In the first configuration ($\text{ReLU-KAN}^1$), parameters $S$ and $E$ are held constant during training. In contrast, the second configuration ($\text{ReLU-KAN}^2$) allows for the learning of $S$ and $E$, making them trainable parameters.


\begin{table}[]
    \centering
    \caption{Training Speed Comparison (unit: second)} \label{tab:tab2}
    \begin{tabular}{c|c|c|c|c|c|c}
    \hline
    Func. & KAN & KAN & $\text{ReLU-KAN}^1$ & $\text{ReLU-KAN}^1$ & $\text{ReLU-KAN}^2$ & $\text{ReLU-KAN}^2$ \\ 
         & (CPU) & (GPU) & (CPU) & (GPU) & (CPU) & (GPU) \\ \hline
    $f_1$ & 2.80 & 6.01 & 0.56 & 0.72 & 0.56 & 0.73\\ \hline
    $f_2$ & 3.04 & 7.23 & 0.78 & 0.75 & 0.80 & 0.72 \\ \hline
    $f_3$ & 5.30 & 12.70 & 1.21 & 1.13 & 1.25 & 1.05 \\ \hline
    $f_4$ & 11.30 & 23.12 & 1.57 & 1.08 & 1.59 & 1.15 \\ \hline
    $f_5$ & 19.23 & 34.38 & 2.26 & 1.15 & 2.32 & 1.10 \\ \hline
    \end{tabular}
\end{table}

Based on the results presented in the Table~\ref{tab:tab2}, the following conclusions can be drawn:

\begin{itemize}
    \item \textbf{ReLU-KAN is faster than KAN:} ReLU-KAN is significantly less time-consuming than KAN in all comparisons.
    \item \textbf{ReLU-KAN training scales more efficiently with complexity:} As the model architecture becomes more complex, the training time increases for both KAN and ReLU-KAN. However, the improvement of time consumption of ReLU-KAN is much smaller than that of KAN
    \item \textbf{ReLU-KAN's GPU speed advantage grows with model complexity:} ReLU-KAN demonstrates a more significant speed advantage on GPU compared to CPU as the model complexity increases. For a single-layer model ($f_1$ and $f_2$), ReLU-KAN is 4 times faster than KAN. For a 2-layer model ($f_3$ and $f_4$), the speed difference ranges from 5 to 10 times, and for a 3-layer model($f_5$), the speed difference approaches 20 times.
    \item \textbf{Learning of $S$ and $E$ is free in time:} Whether $S$ and $E$ are set to be trainable or not has almost no effect on the time consumption.
\end{itemize}

\subsection{Comparison of Fitting Ability}
The fitting capabilities of KAN and ReLU-KAN are then compared on three unary functions and three multivariate functions, each of which uses the parameter Settings shown in the Table~\ref{tab:tab3}.

\begin{table}[]
    \centering
    \caption{Parameter Settings for Fitting Ability Comparison Experiments} \label{tab:tab3}
    \begin{tabular}{l|l|l}
    \hline
    Func. & KAN Setting & ReLU-KAN Setting \\ \hline
    $f_1: \text{sin}(\pi x)$ 
    & width=[1, 1], G=5, k=3 & width=[1, 1], G=5, k=3 \\ \hline
    $f_2: sin(5\pi x) + x$ 
    & width=[2, 1], G=5, k=3 & width=[2, 1], G=5, k=3 \\ \hline
    $f_3: e^x$
    & width=[2, 1, 1], G=5, k=3 & width=[2, 1, 1], G=5, k=3 \\ \hline
    $f_4: \text{sin}(\pi x_1 + \pi x_2)$
    & width=[2, 5, 1], G=5, k=3 & width=[2, 5, 1], G=5, k=3 \\ \hline
    $f_5: e^{\text{sin}(\pi x_1) + x_2^2}$
    & width=[2, 5, 1], G=5, k=3 & width=[2, 5, 1], G=5, k=3 \\ \hline
    $f_6: e^{\text{sin}(\pi x_1^2 + \pi x_2^2)+\text{sin}(\pi x_3^2 + \pi x_4^2)}$
    & width=[4, 4,  2, 1], G=10, k=3 & width=[4, 4, 2, 1], G=10, k=3 \\ \hline
\end{tabular}
\end{table}

There are also two configurations of ReLU-KAN in this experiment. In the first configuration (ReLU-KAN$^1$), the parameters $S$ and $E$ are kept constant during training. By contrast, The second configuration (ReLU-KAN$^2$) allows learning $S$ and $E$.

To assess the performance of KAN and ReLU-KAN, we employed the Mean Squared Error (MSE) loss function as the evaluation metric and utilized the Adam optimizer for optimization. The maximum number of iterations was set to 1000.

To visualize the iterative process of both models, we  plotted their loss curves. And we can visualize the fit in the following way: for univariate functions $f_1, f_2$, and $f_3$, we directly plotted their original $f(x)$ curves alongside the fitted curves, providing a clear visual representation of their fitting performance. For multivariate functions $f_4, f_5$, and $f_6$, we generated scatter plots of predicted values versus true values. The closer the scatter points lie to the line $pred = true$, the better the fitting performance.
\begin{landscape}
    \begin{table}[]
    \centering
    \caption{Fitting Process and Effects}
    \label{tab:tb41}
    \begin{tabular}{c|c|c|c|c}
    \hline
    Func. & Loss curve & KAN & ReLU-KAN$^1$ & ReLU-KAN$^2$ \\ \hline
        $f_1$ &
        \includegraphics[width=0.3\textwidth]{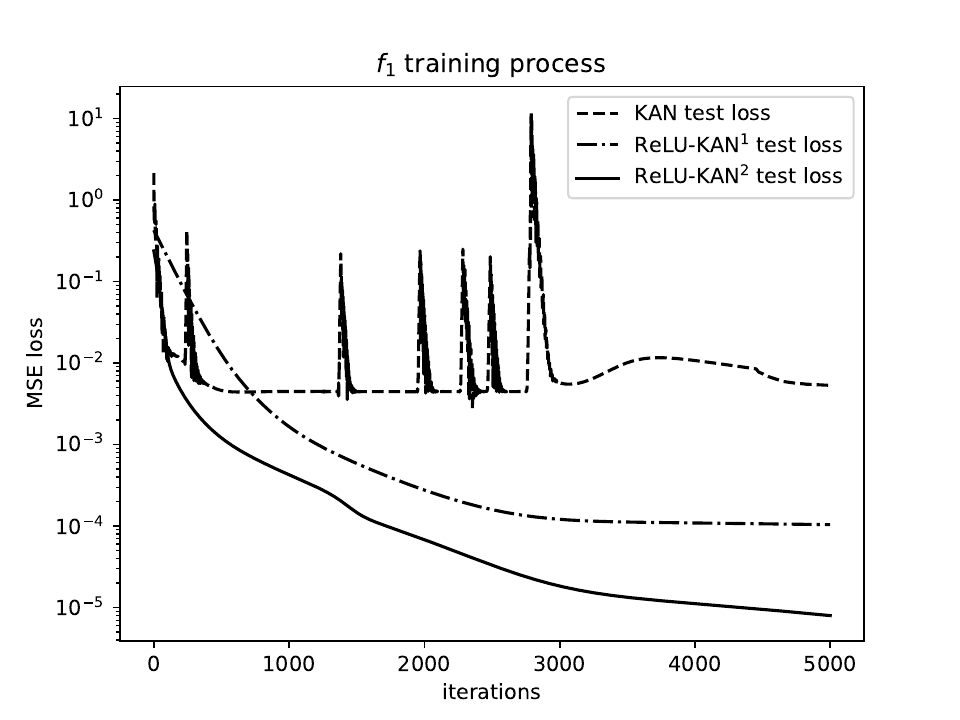} & 
        \includegraphics[width=0.3\textwidth]{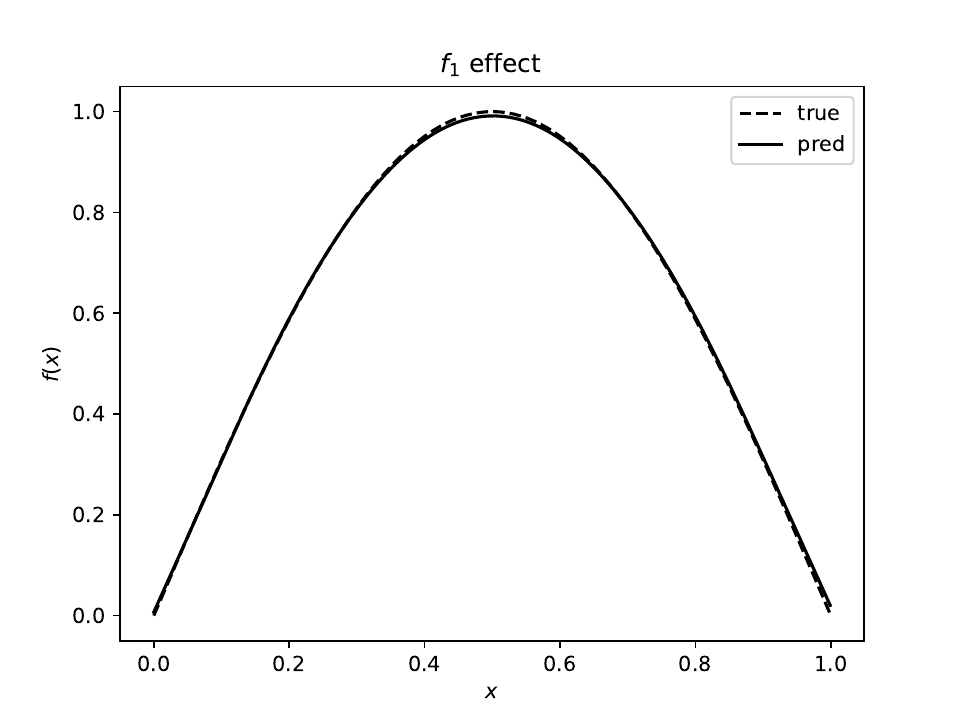} & 
        \includegraphics[width=0.3\textwidth]{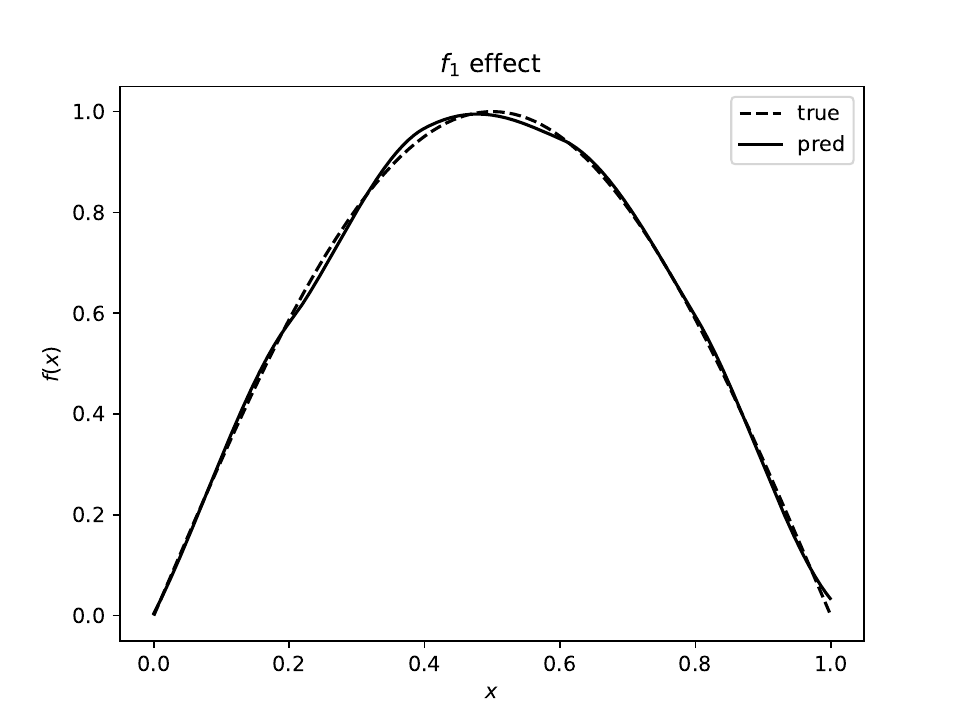}
        & 
        \includegraphics[width=0.3\textwidth]{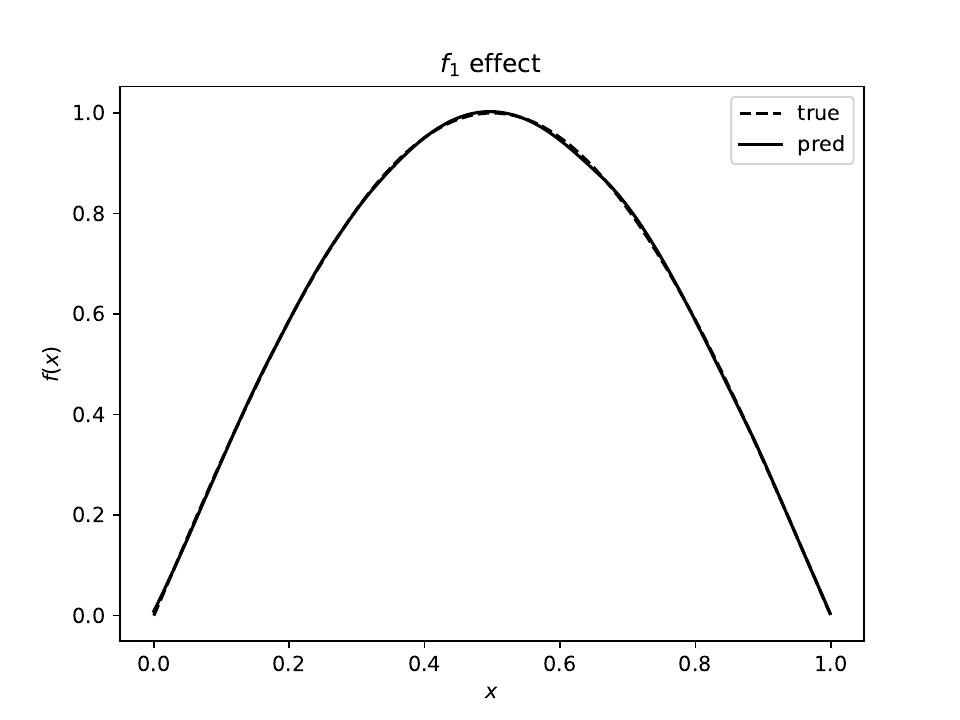} \\ \hline
        $f_2$ &
        \includegraphics[width=0.3\textwidth]{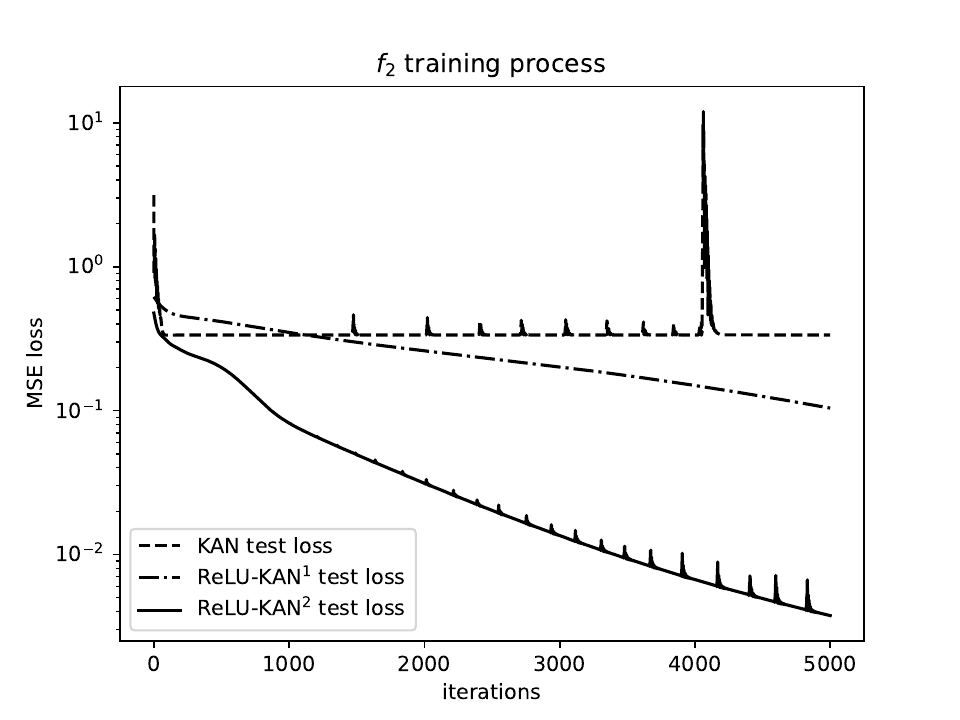} & 
        \includegraphics[width=0.3\textwidth]{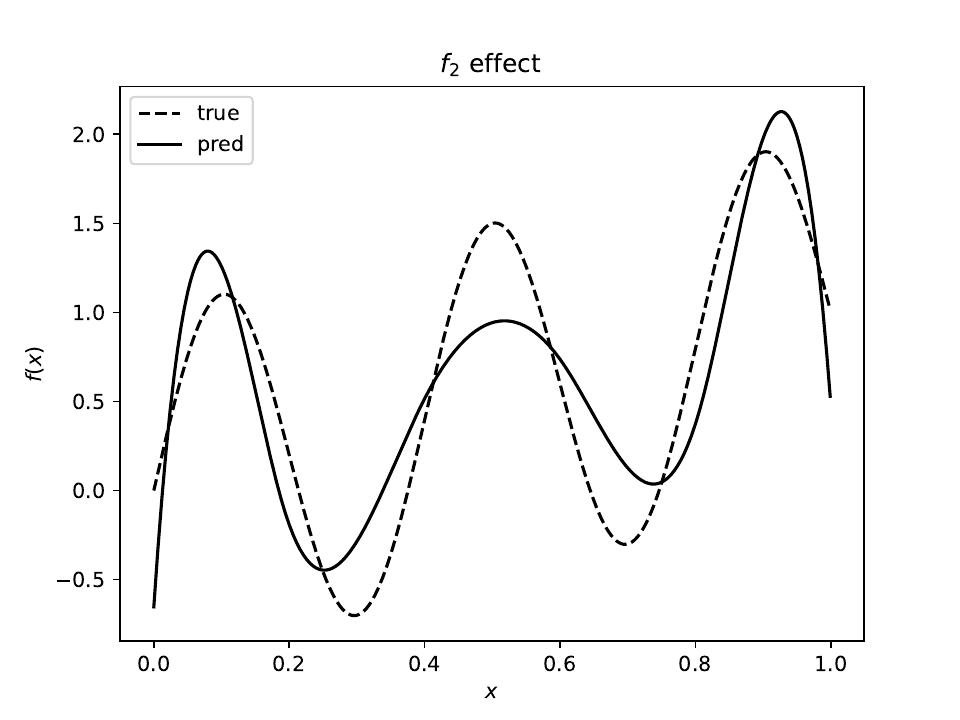} & 
        \includegraphics[width=0.3\textwidth]{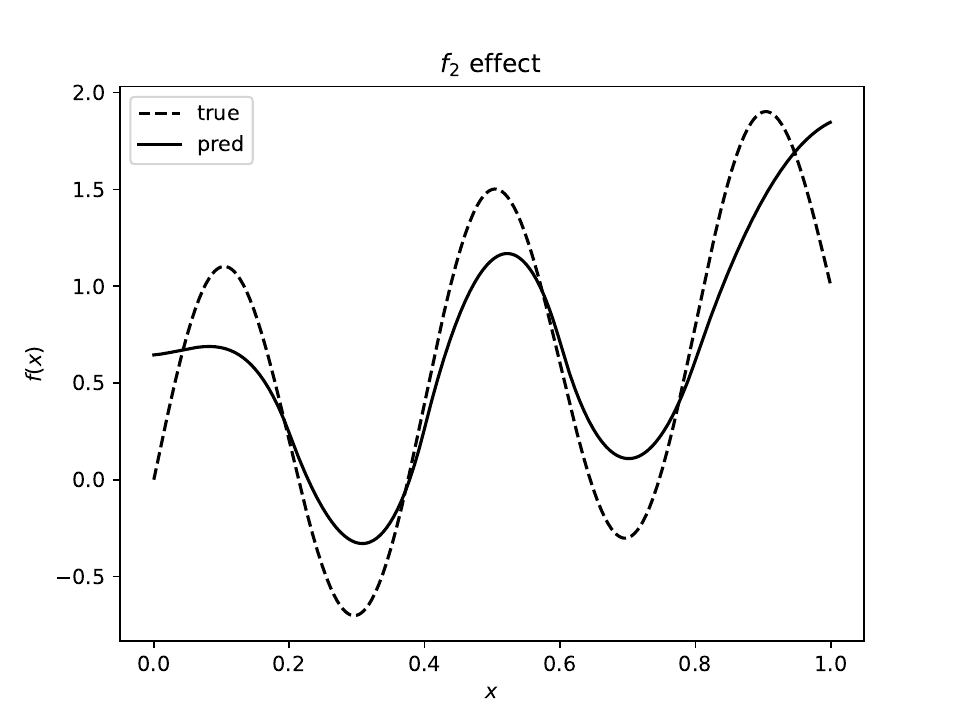}
        & 
        \includegraphics[width=0.3\textwidth]{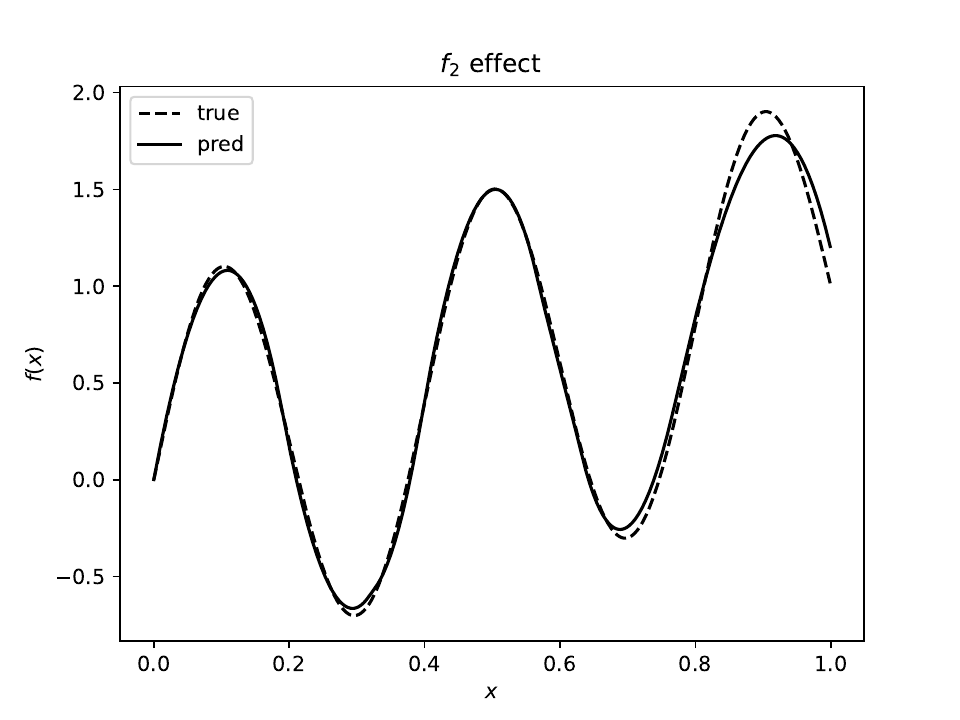} \\ \hline
        $f_3$ &
        \includegraphics[width=0.3\textwidth]{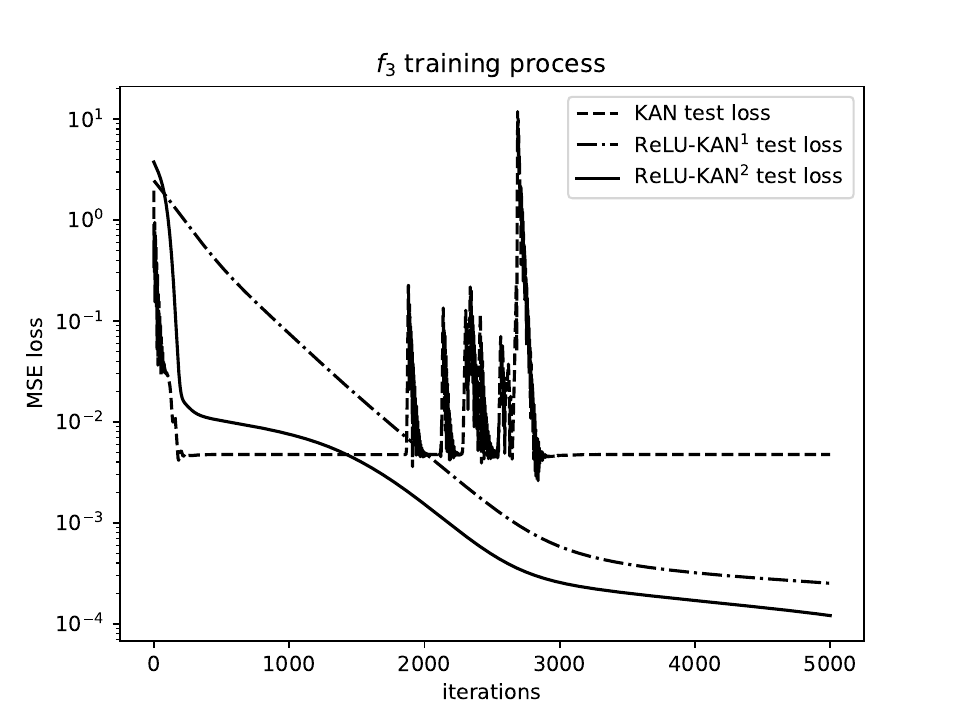} & 
        \includegraphics[width=0.3\textwidth]{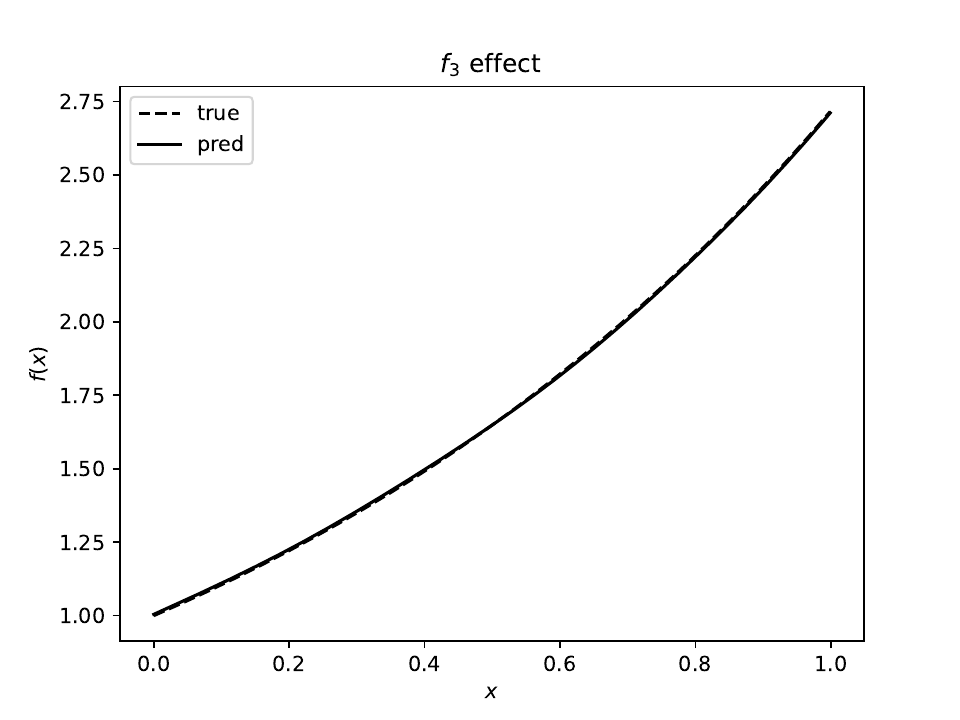} & 
        \includegraphics[width=0.3\textwidth]{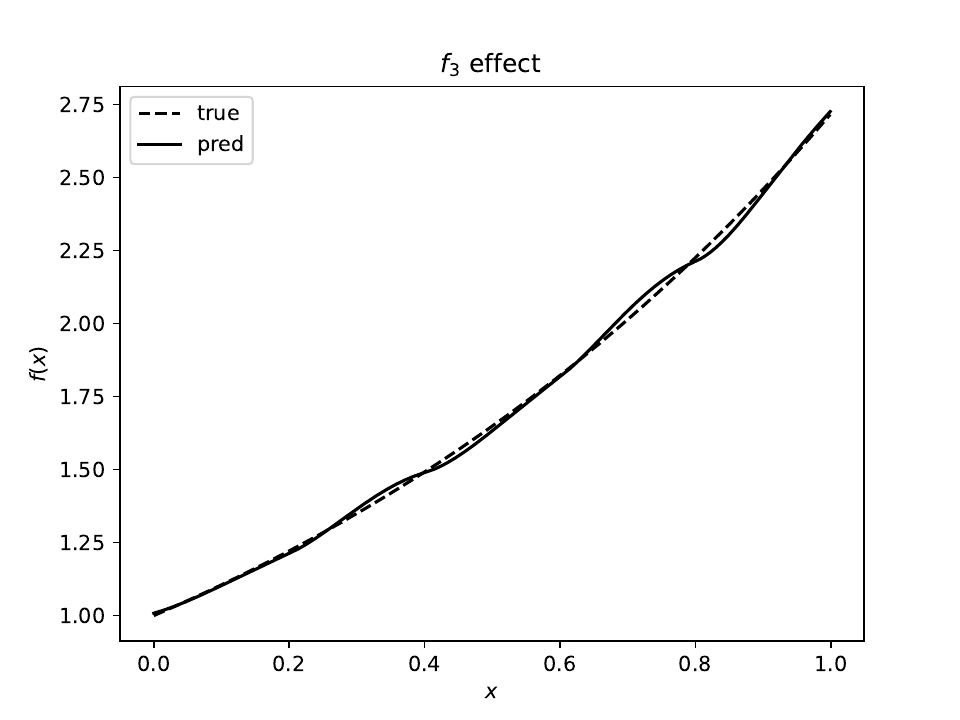}& 
        \includegraphics[width=0.3\textwidth]{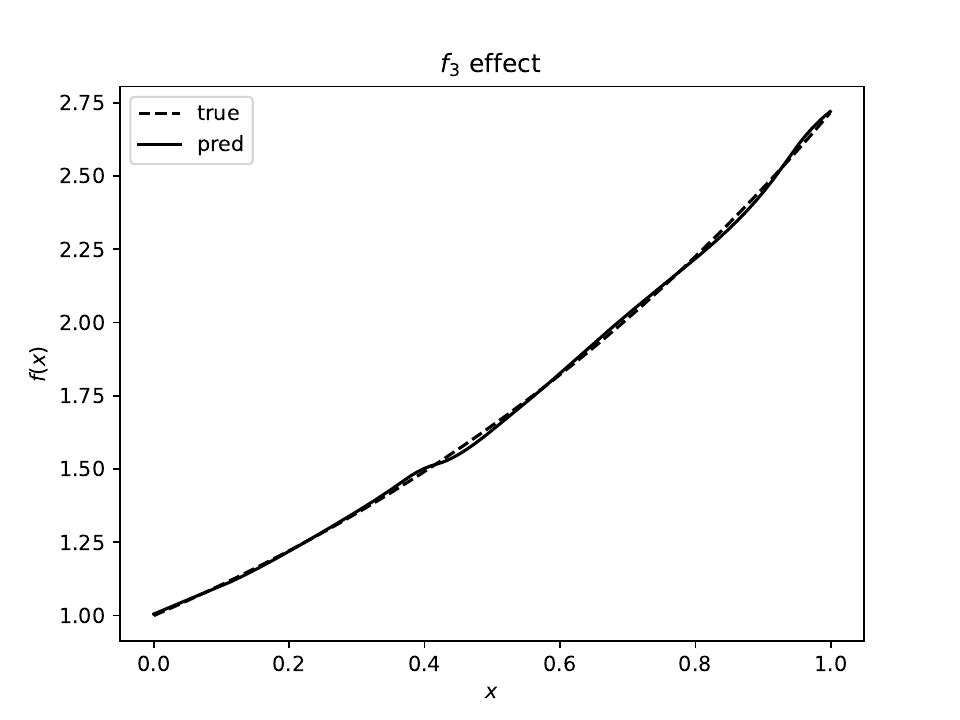} \\ \hline
    \end{tabular}
\end{table}

\end{landscape}

\begin{landscape}

\begin{table}[]
    \centering
    \caption{Fitting Process and Effects}
    \label{tab:tb42}
    \begin{tabular}{c|c|c|c|c}
    \hline
    Func. & Loss curve & KAN & ReLU-KAN$^1$ & ReLU-KAN$^2$ \\ \hline
        $f_4$ &
        \includegraphics[width=0.3\textwidth]{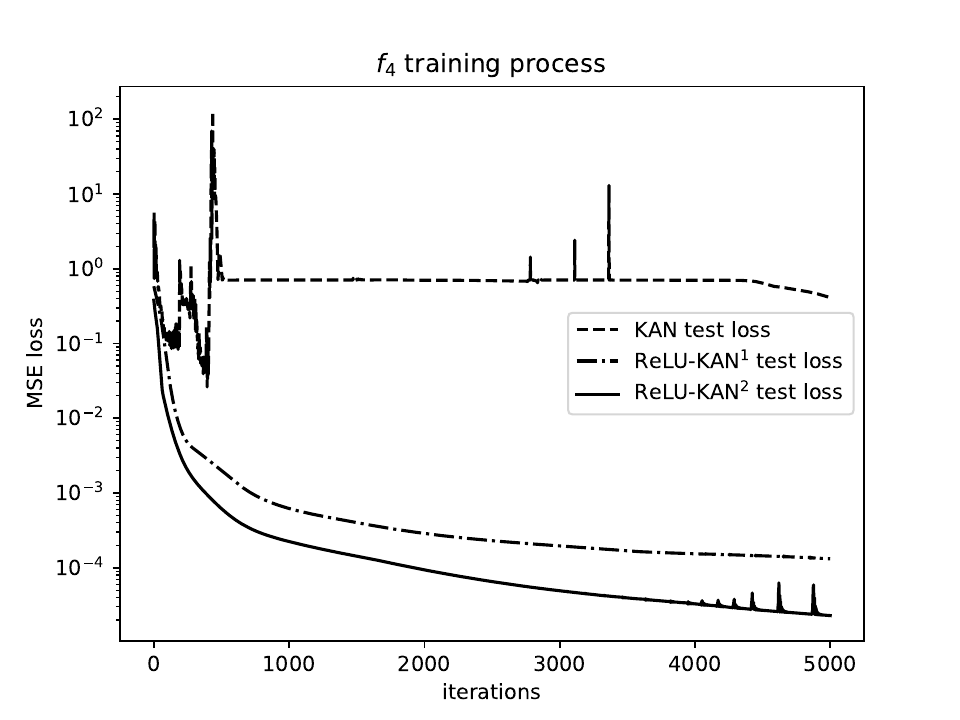} & 
        \includegraphics[width=0.3\textwidth]{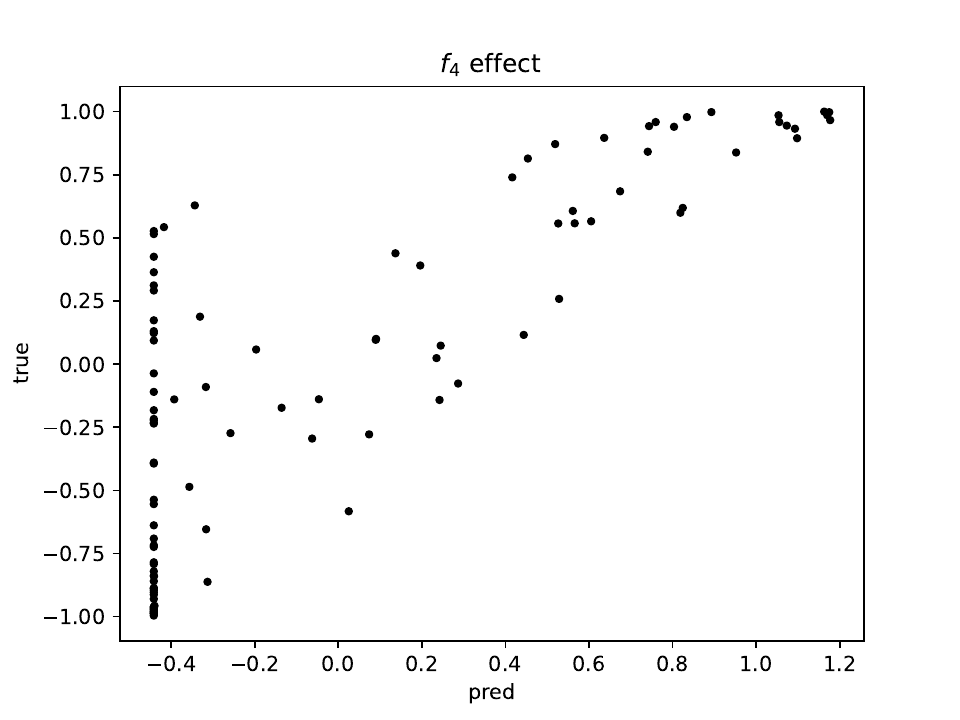} & 
        \includegraphics[width=0.3\textwidth]{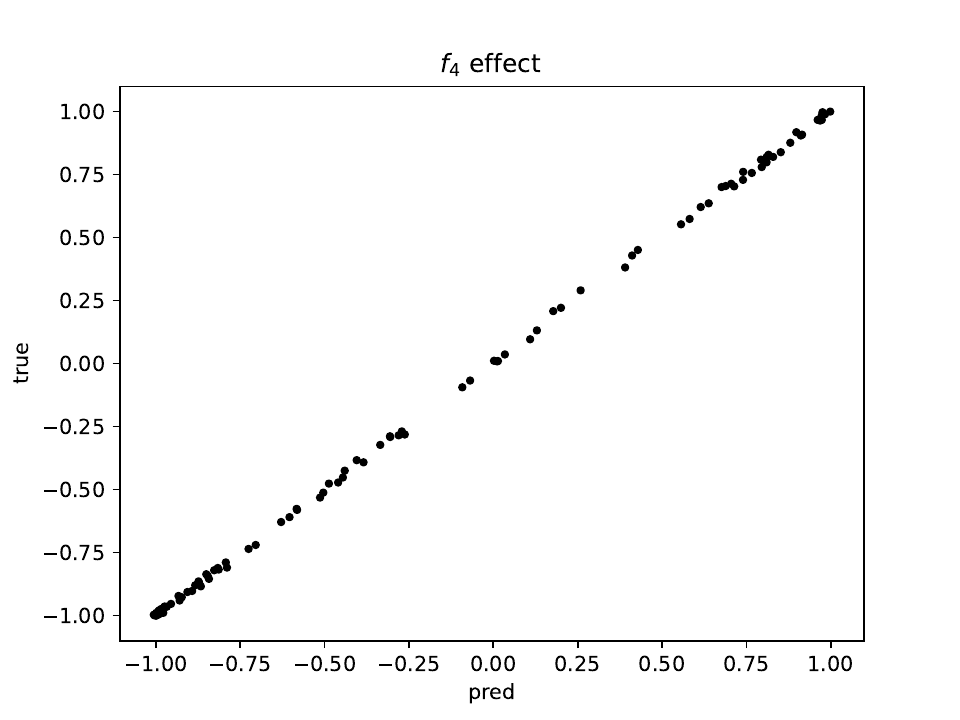}& 
        \includegraphics[width=0.3\textwidth]{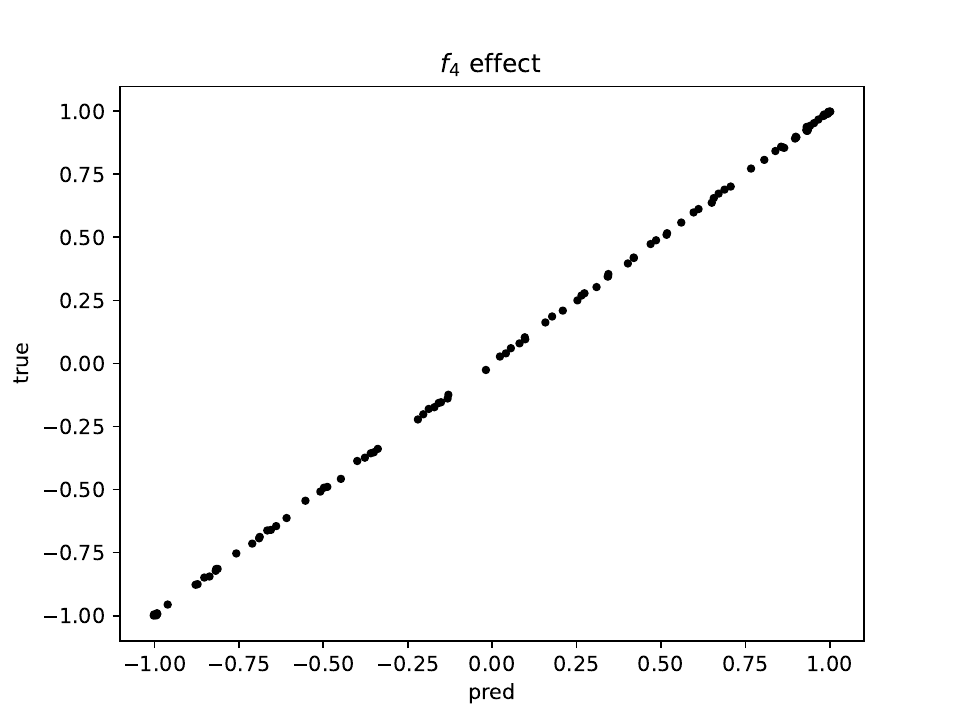} \\ \hline
        $f_5$ &
        \includegraphics[width=0.3\textwidth]{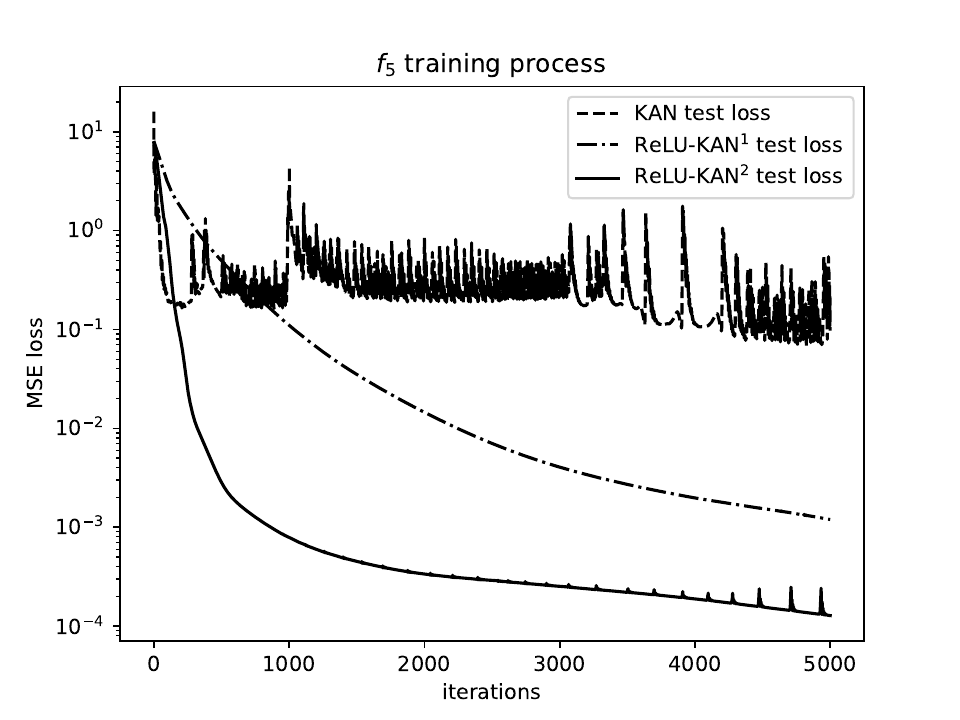} & 
        \includegraphics[width=0.3\textwidth]{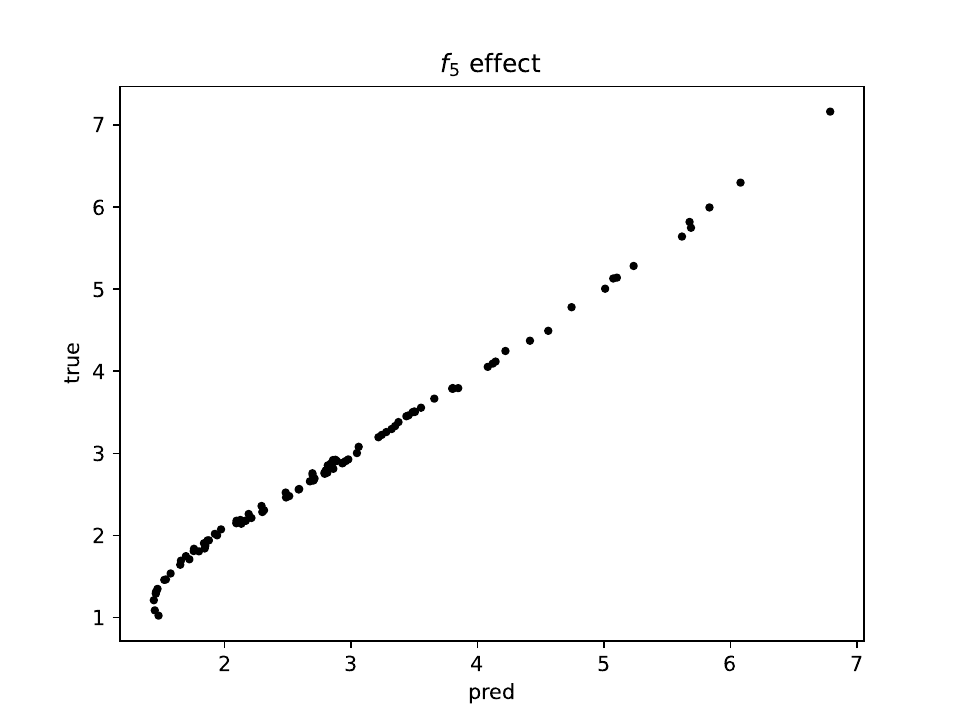} & 
        \includegraphics[width=0.3\textwidth]{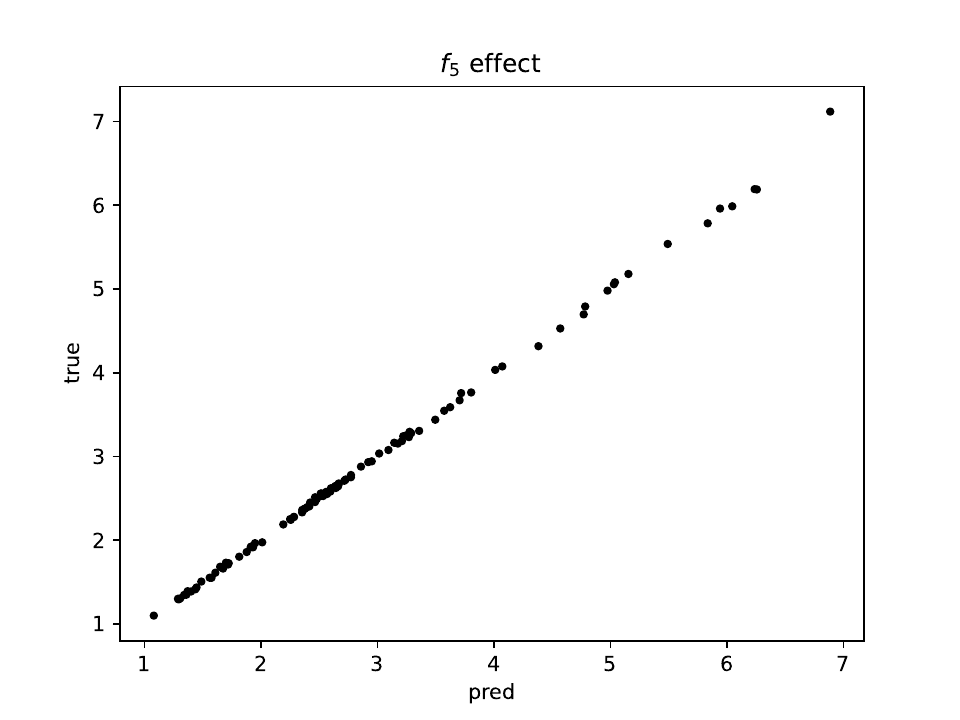}& 
        \includegraphics[width=0.3\textwidth]{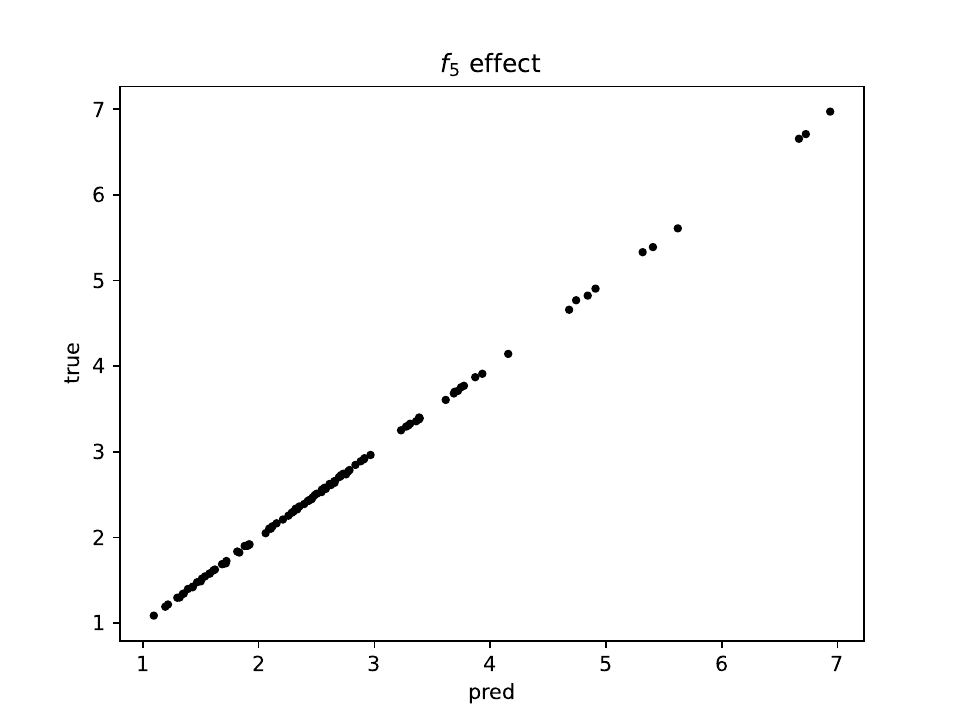} \\ \hline
        $f_6$ &
        \includegraphics[width=0.3\textwidth]{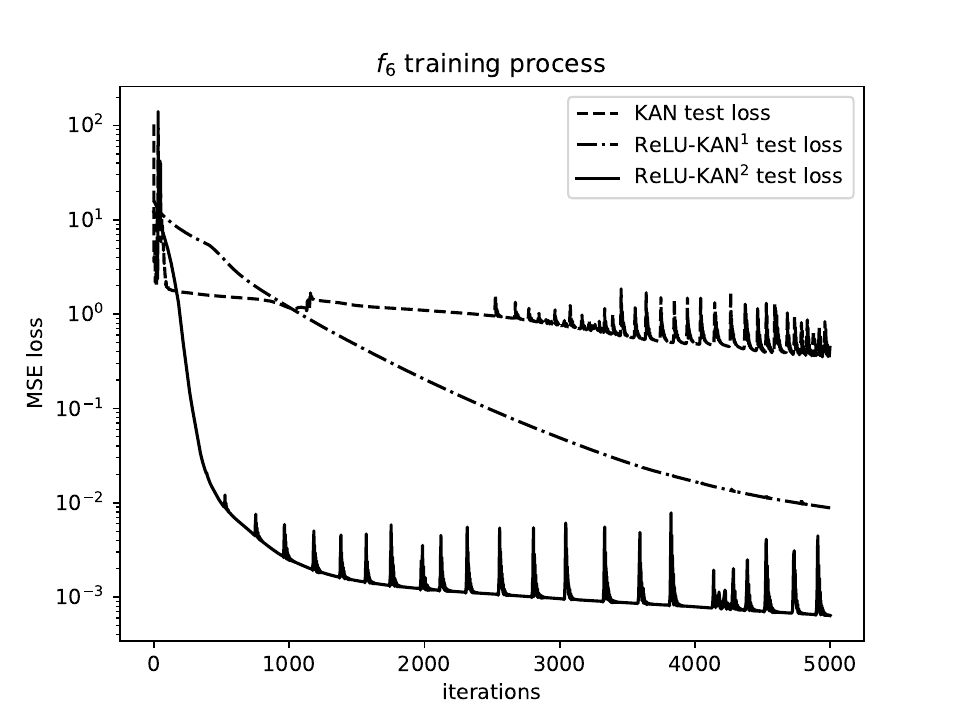} & 
        \includegraphics[width=0.3\textwidth]{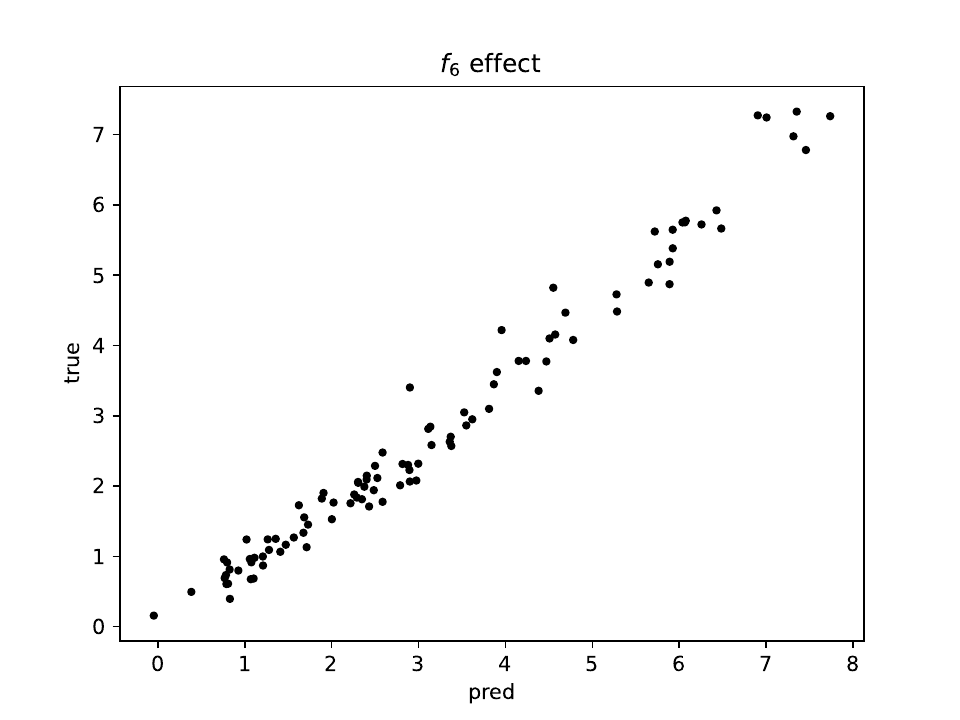} & 
        \includegraphics[width=0.3\textwidth]{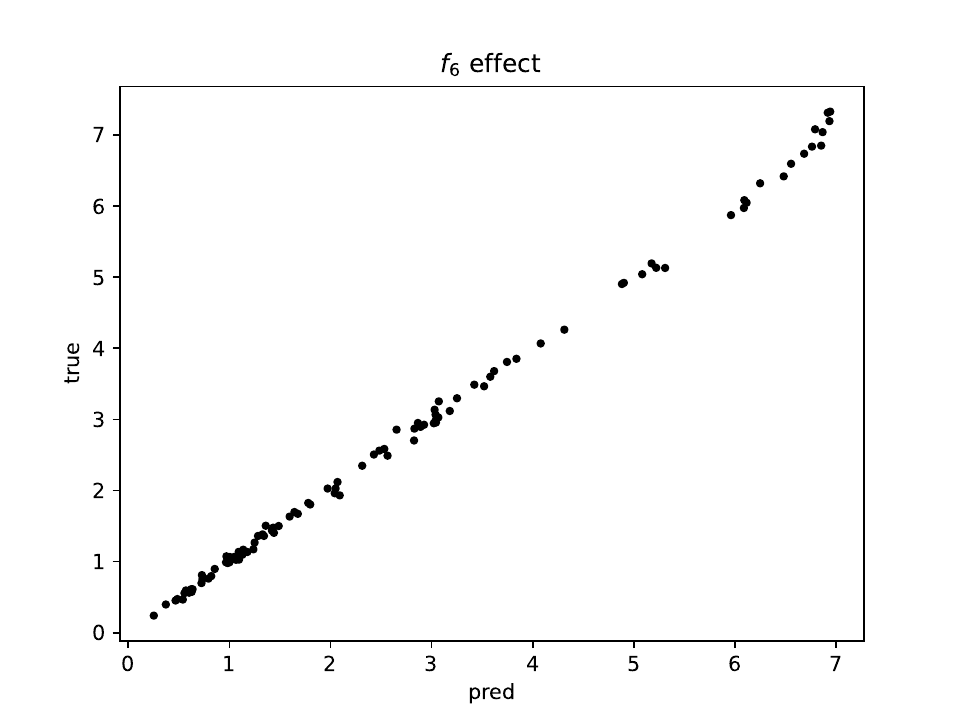}& 
        \includegraphics[width=0.3\textwidth]{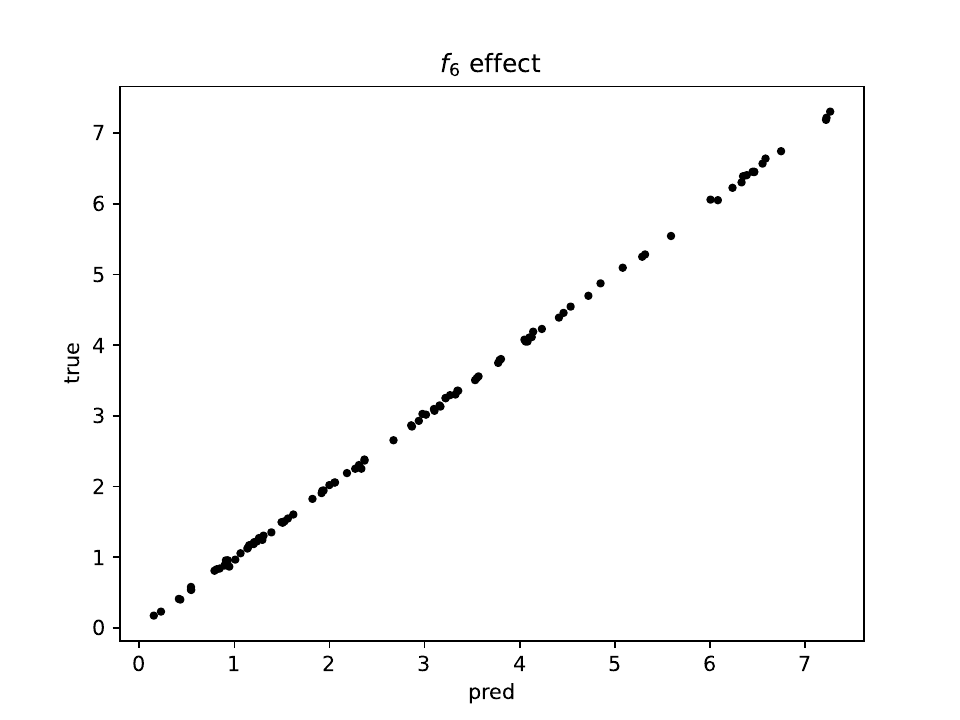} \\ \hline
    \end{tabular}
\end{table}

\end{landscape}

\begin{table}[]
    \centering
    \caption{Training Speed Comparison} \label{tab:tab5}
    \begin{tabular}{c|c|c|c}
    \hline
    Func. & KAN & $\text{ReLU-KAN}^1$ & $\text{ReLU-KAN}^2$ \\ \hline
    $f_1$ & $2.72 \times 10^{-3}$ & $1.04 \times 10^{-4}$ & $7.96 \times 10^{-6}$ \\ \hline
    $f_2$ & $3.34 \times 10^{-1}$ & $1.04 \times 10^{-1}$ & $7.96 \times 10^{-3}$  \\ \hline
    $f_3$ & $2.62 \times 10^{-3}$ & $2.52 \times 10^{-4}$ & $1.20 \times 10^{-4}$ \\ \hline
    $f_4$ & $2.62 \times 10^{-2}$ & $1.32 \times 10^{-4}$ & $2.29 \times 10^{-5}$ \\ \hline
    $f_5$ & $6.80 \times 10^{-2}$ & $1.20 \times 10^{-3}$ & $1.28 \times 10^{-4}$ \\ \hline
    $f_6$ & $3.47 \times 10^{-1}$ & $8.83 \times 10^{-3}$ & $6.37 \times 10^{-4}$ \\ \hline
\end{tabular}
\end{table}

The results in the Table~\ref{tab:tb41} and Table~\ref{tab:tb42} indicate that ReLU-KAN exhibits a more stable training process and achieves higher fitting accuracy compared to the KAN, given identical network structure and scale. This advantage becomes particularly pronounced for multi-layer networks, especially when fitting functions like $f_2$ with a higher frequency of change. In these cases, ReLU-KAN demonstrates superior fitting capabilities.

In order to more clearly see the accuracy advantage of ReLU-KAN compared with KAN, the results of Table~\ref{tab:tb41} and Table~\ref{tab:tb42} on the test set are statistically analyzed in Table~\ref{tab:tab5} using the scientific and technical method, from which it can be seen that ReLU-KAN$^2$ accuracy can be 1-3 orders of magnitude higher than that of KAN. Comparing ReLU-KAN$^1$ and ReLU-KAN$^2$, we can find that although ReLU-KAN$^1$ is also better than KAN, it is not as obvious as ReLU-KAN$^2$. This can confirm the necessity to endow the basis functions with plasticity.

\begin{table}[]
    \centering
    \caption{Catastrophic Forgetting Experiments}
    \label{tab:tab5}
    \begin{tabular}{c|c|c|c|c|c}
    \hline
    & Phase 1 & Phase 2 & Phase 3 & Phase 4 & Phase 5 \\ \hline
        input &
        \includegraphics[width=0.15\textwidth]{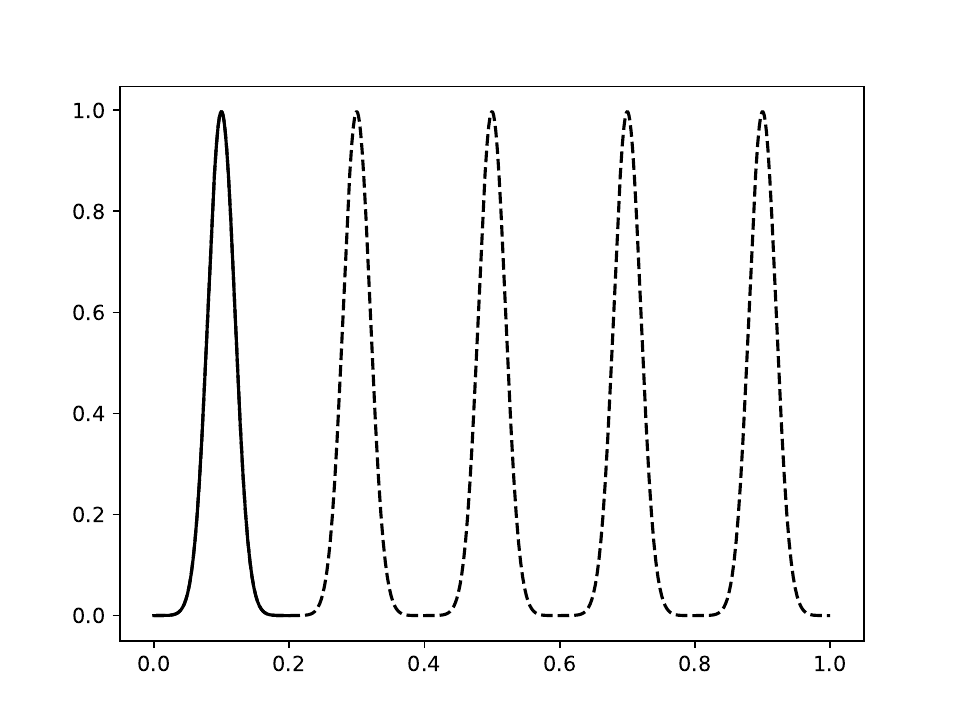} & 
        \includegraphics[width=0.15\textwidth]{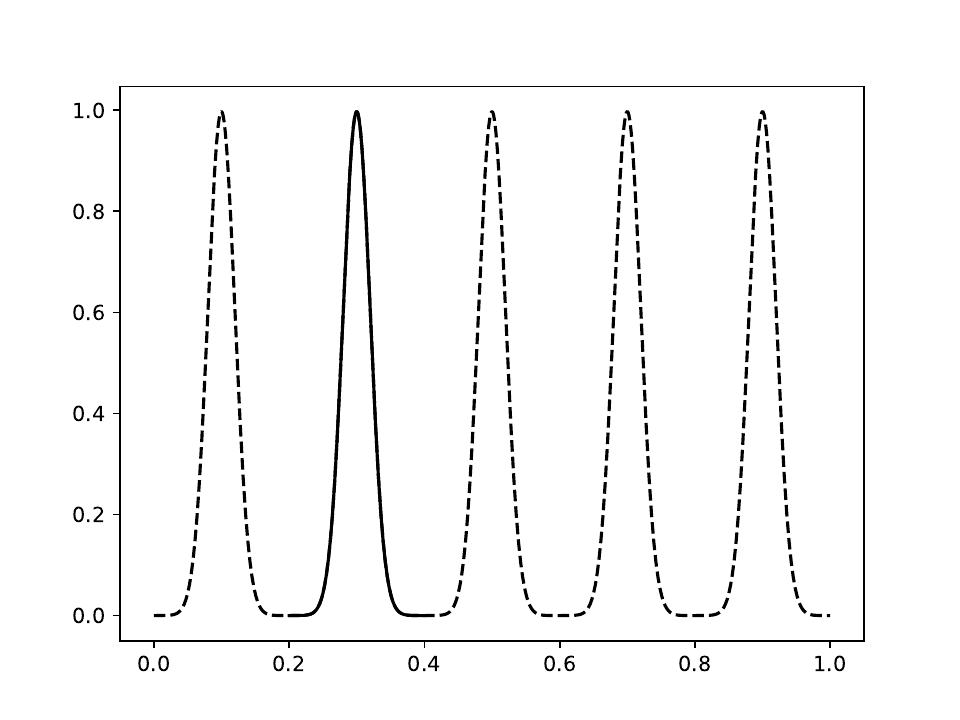} & 
        \includegraphics[width=0.15\textwidth]{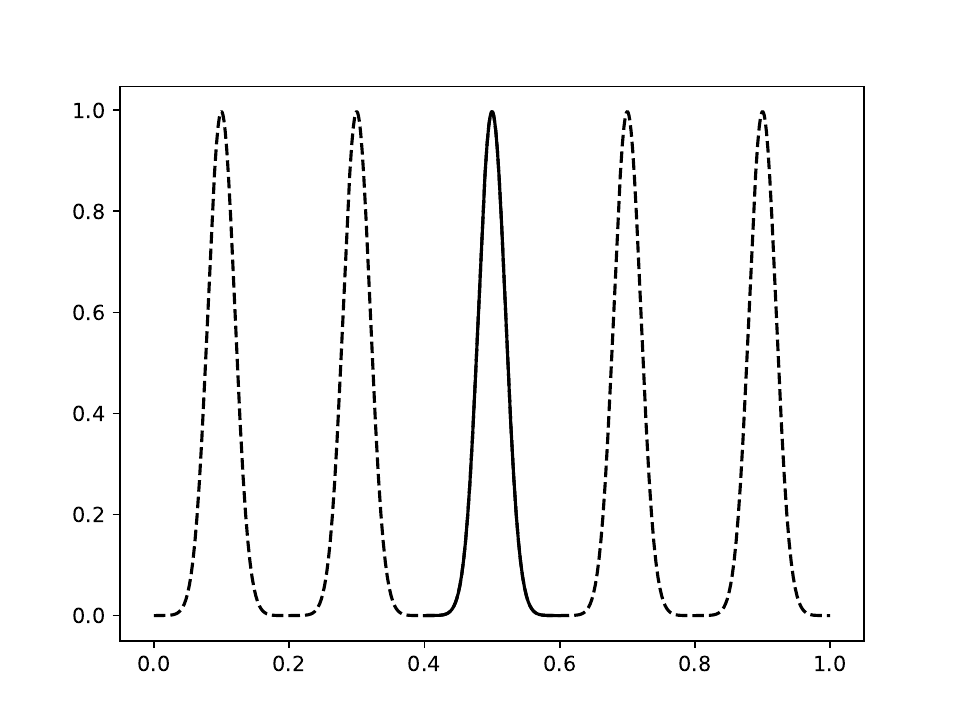} &
        \includegraphics[width=0.15\textwidth]{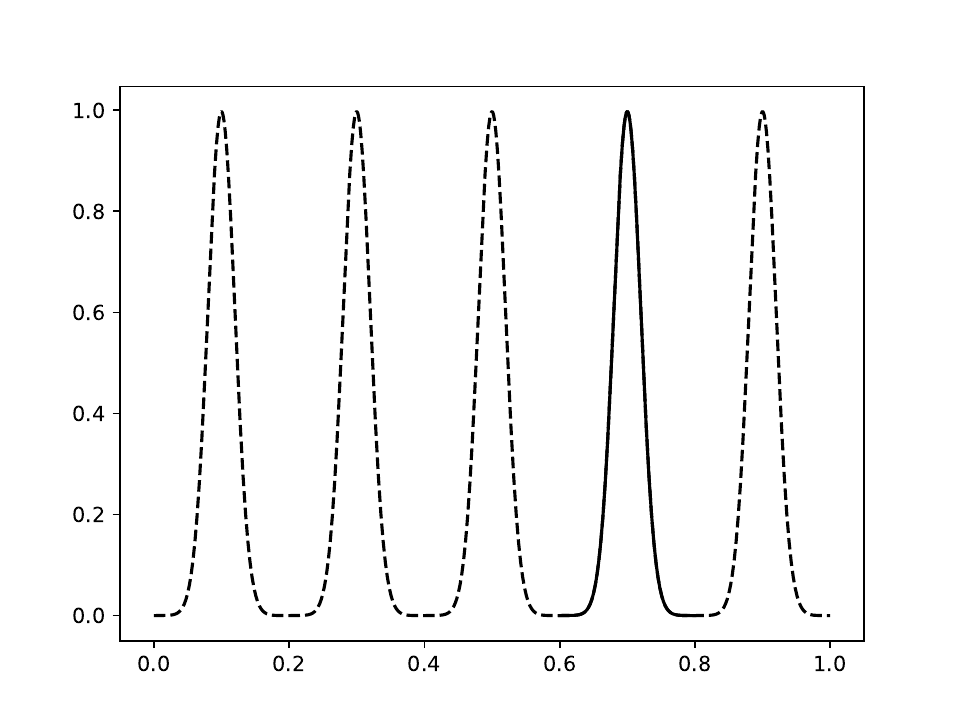} & 
        \includegraphics[width=0.15\textwidth]{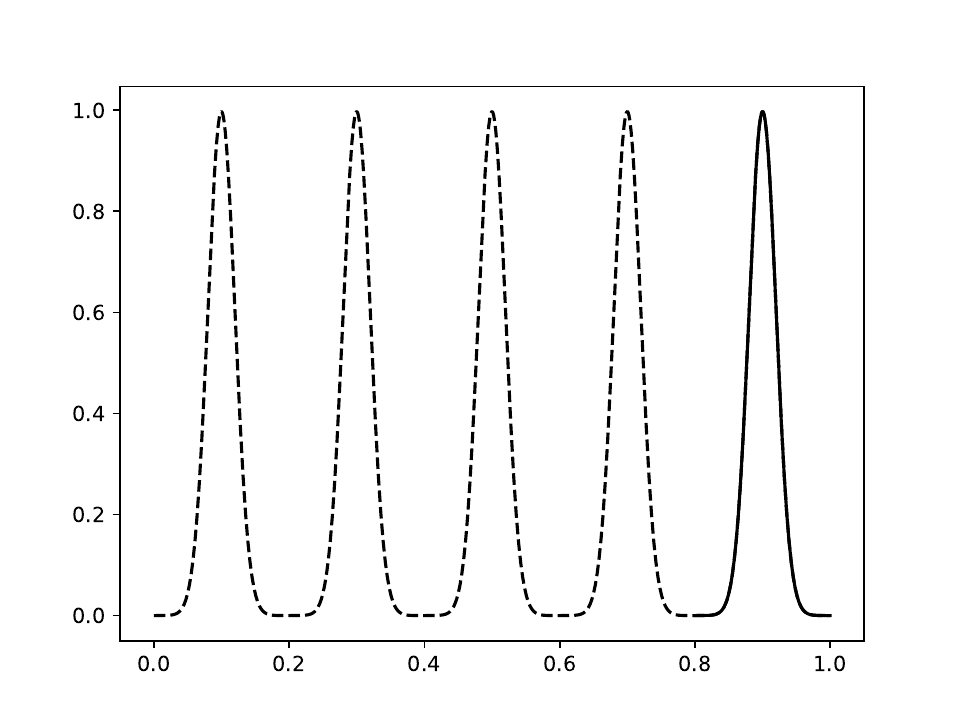}\\ \hline
        output &
        \includegraphics[width=0.15\textwidth]{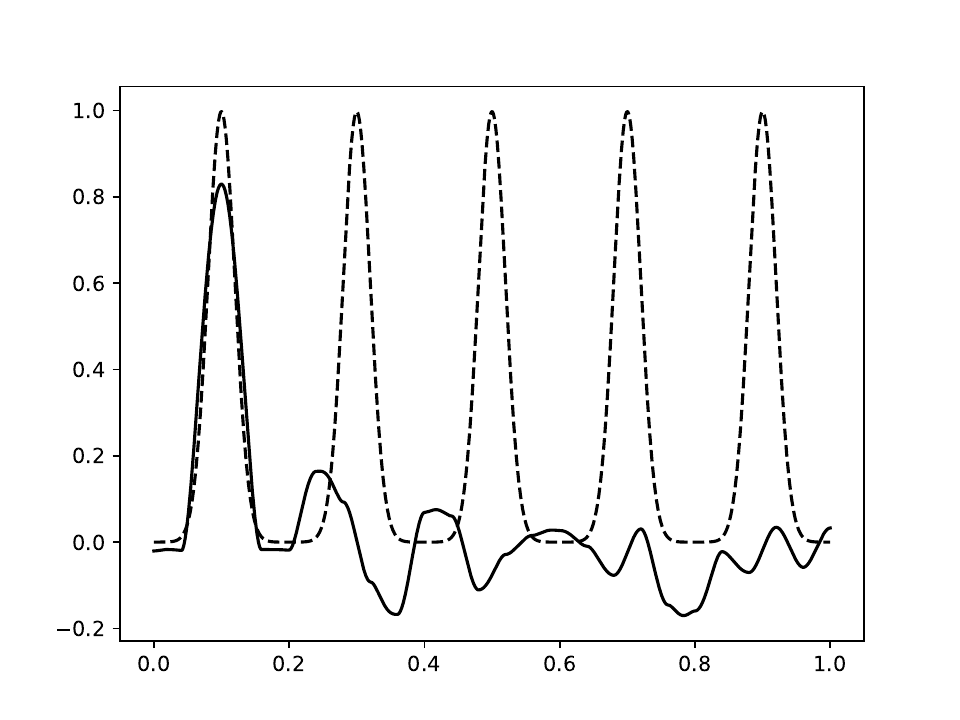} & 
        \includegraphics[width=0.15\textwidth]{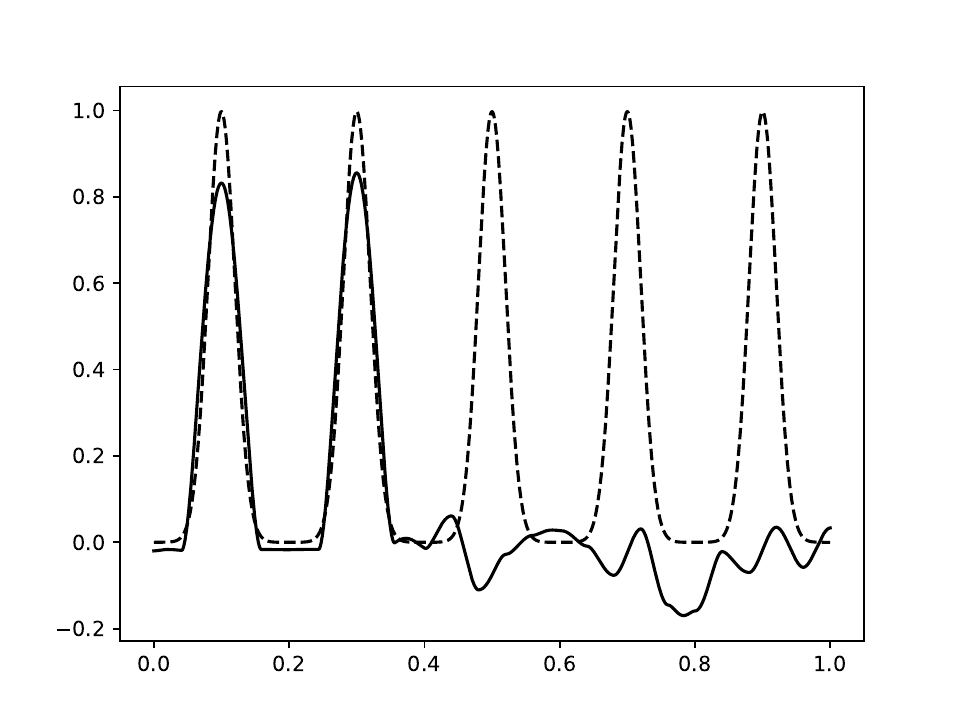} & 
        \includegraphics[width=0.15\textwidth]{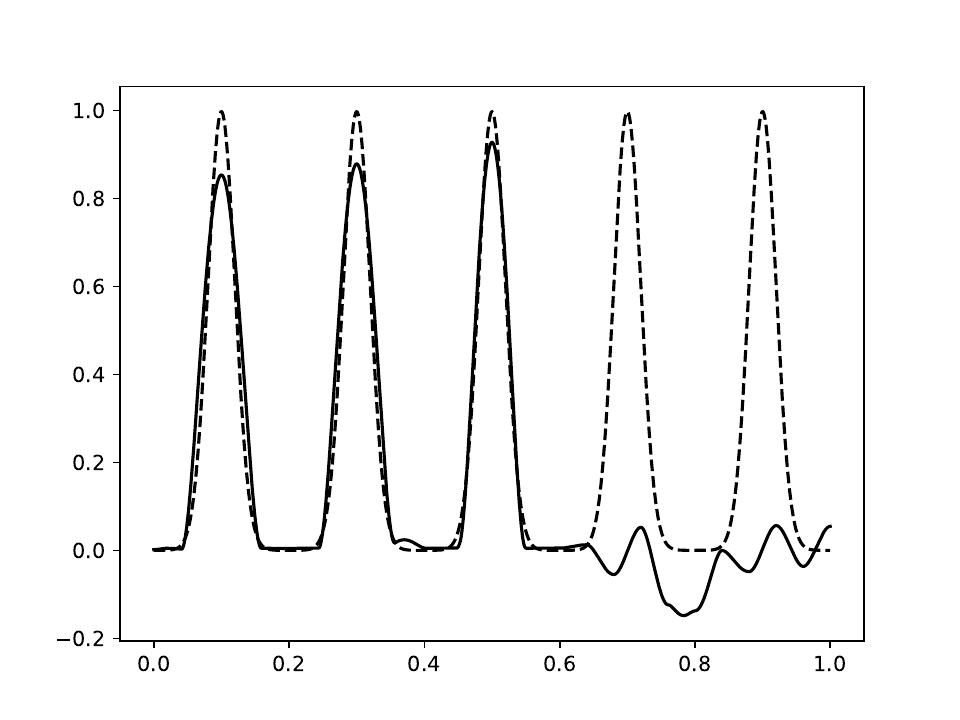} &
        \includegraphics[width=0.15\textwidth]{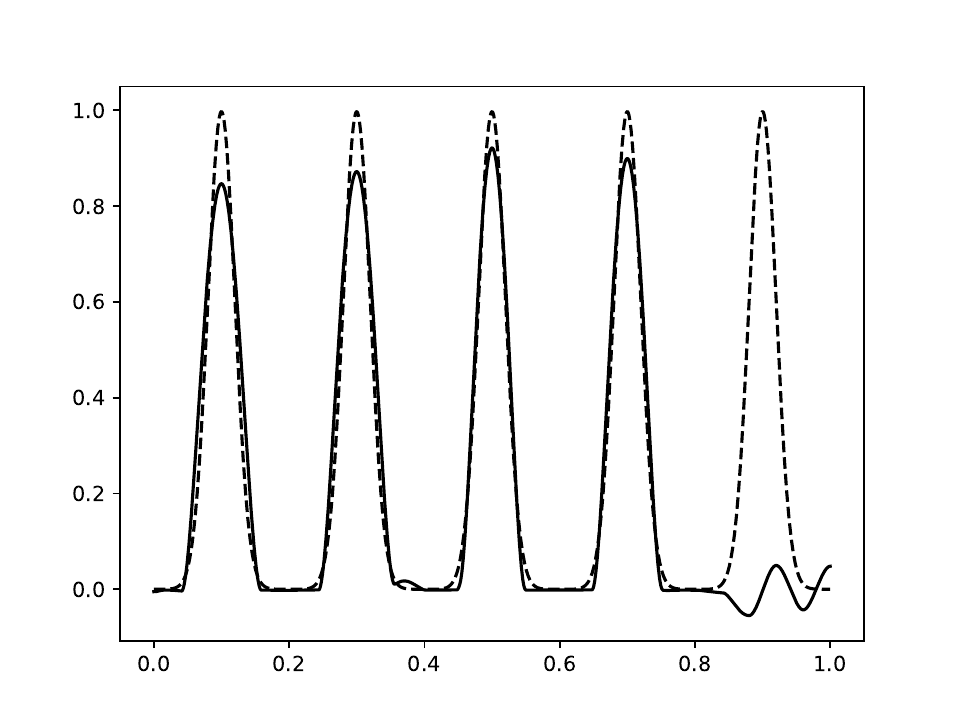} & 
        \includegraphics[width=0.15\textwidth]{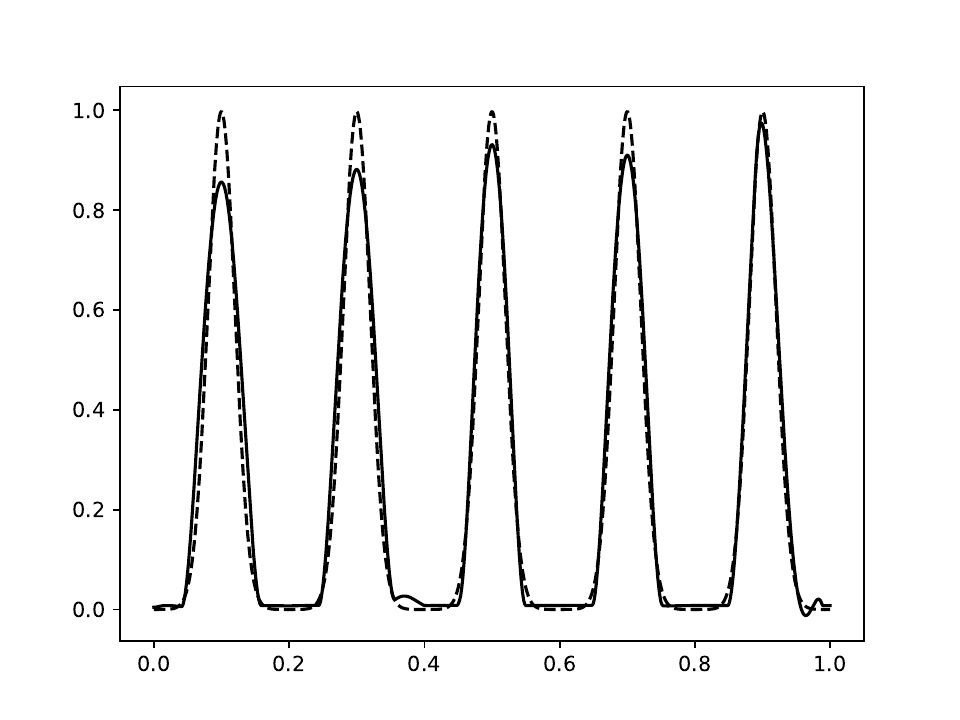}\\ \hline
    \end{tabular}
\end{table}

\subsection{ReLU-KAN Avoids Catastrophic Forgetting}

Leveraging its similar basis function structure to KAN, ReLU-KAN is expected to inherit KAN's resistance to catastrophic forgetting. To verify this, we conducted a simple experiment.

Similar to the experiment designed for KAN, the target function has five peaks. During training, the model is presented with data for only one peak at a time. The following figure illustrates the fitting curve of ReLU-KAN after each training iteration.

As shown in the Table~\ref{tab:tab5}, ReLU-KAN similarly has the ability to avoid catastrophic forgetting.

\section{Summary and Prospect}
This paper introduces ReLU-KAN, a novel architecture that significantly outperforms conventional KANs by replacing B-splines with simplified, trainable basis functions. ReLU-KAN achieves up to 20x faster training and one to three orders of magnitude accuracy while maintaining the ability to prevent catastrophic forgetting. Unlike B-splines, the proposed basis functions only consist of matrix operations and ReLU activations, which achieve efficient GPU parallelization and powerful fitting capabilities. Experimental results show that ReLU-KAN effectively balances computational efficiency and model expression ability, and simple basis functions can meet the approximation of complex functions. Future research will explore alternative basis functions in the KAN framework to further optimize the performance.


\end{document}